\pdfoutput=1
\documentclass[lettersize,journal]{IEEEtran}
\usepackage{amsmath,amsfonts}
\usepackage{algorithmic}
\usepackage{algorithm}
\usepackage{array}
\usepackage[caption=false,font=normalsize,labelfont=sf,textfont=sf]{subfig}
\usepackage{textcomp}
\usepackage{stfloats}
\usepackage{url}
\usepackage{verbatim}
\usepackage{graphicx}
\usepackage{cite}
\usepackage{multirow}
\usepackage{caption}
\usepackage{colortbl}
\def\BibTeX{{\rm B\kern-.05em{\sc i\kern-.025em b}\kern-.08em
    T\kern-.1667em\lower.7ex\hbox{E}\kern-.125emX}}
\usepackage{balance}

\begin{document}
% \title{How to Use the IEEEtran \LaTeX \ Templates}
\title{FlowFace++: Explicit Semantic Flow-supervised End-to-End Face Swapping}

%\author{IEEE Publication Technology Department
% \thanks{Manuscript created October, 2020; This work was developed by the IEEE Publication Technology Department. This work is distributed under the \LaTeX \ Project Public License (LPPL) ( http://www.latex-project.org/ ) version 1.3. A copy of the LPPL, version 1.3, is included in the base \LaTeX \ documentation of all distributions of \LaTeX \ released 2003/12/01 or later. The opinions expressed here are entirely that of the author. No warranty is expressed or implied. User assumes all risk.}
%}

\author{Yu~Zhang,~%\IEEEmembership{Member,~IEEE,}
        ~Hao~Zeng,~%\IEEEmembership{Fellow,~OSA,}
        ~Bowen~Ma,~%\IEEEmembership{Fellow,~OSA,}
        ~Wei~Zhang,~%\IEEEmembership{Fellow,~OSA,}
        ~Zhimeng~Zhang,~%\IEEEmembership{Fellow,~OSA,}
        ~Yu~Ding$^\ast$,~%\IEEEmembership{Life~Fellow,~IEEE}%
        ~Tangjie~Lv,
        ~Changjie~Fan
% \IEEEcompsocitemizethanks{\IEEEcompsocthanksitem J. Chen, C. Fan, Z. Zhang, G. Li, Z. Zhao, J. Bu, and Y. Ding are with the NetEase Inc., China.\protect\\
% Yu Ding is the corresponding author \protect\\
% E-mail: dingyu01@corp.netease.com
% \IEEEcompsocthanksitem Z. Deng is with the Department of Computer Science, University of Houston, Houston, TX, USA 77204-3010. \protect\\
% Email: zdeng4@central.uh.edu
% }% <-this % stops an unwanted space
% \thanks{This manuscript was accepted to IEEE Transactions on Visualization and Computer Graphics in September 2021}
}

% \maketitle
\twocolumn[{%
\renewcommand\twocolumn[1][]{#1}%
\maketitle

\begin{center}
    \centering
    % \begin{figure*}
    \includegraphics[width=1.0\textwidth]{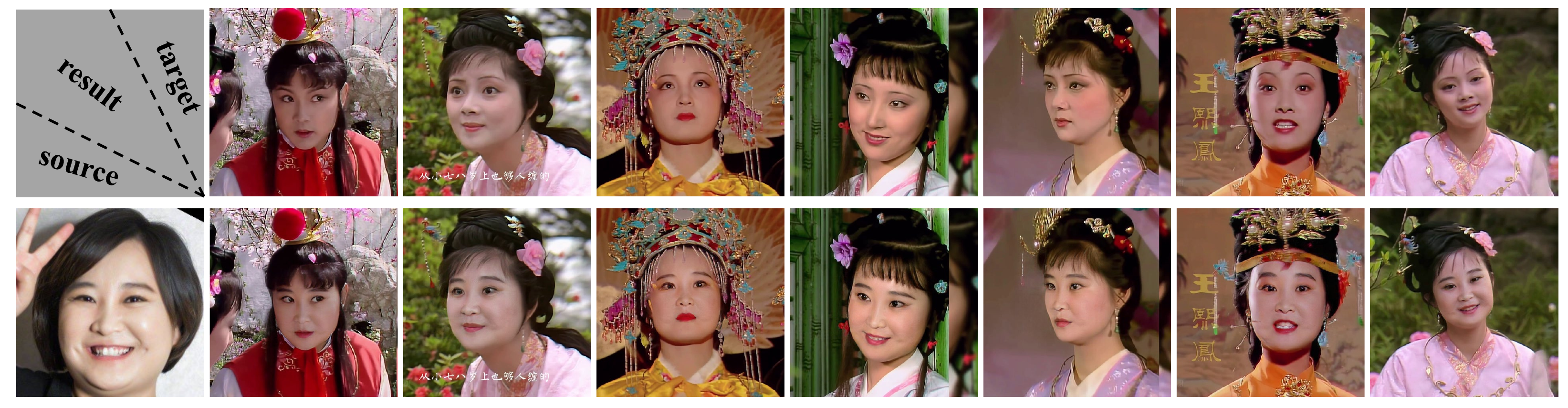}
    \captionof{figure}{Face-swapped images generated by our FlowFace++ model. In the results of swapped images, not only are the inner facial details of the target faces replaced with those of the source faces, but also the facial outlines of the result faces bear similarity to those of the source faces.}
    \label{fig:headline}
    % \end{figure*}
\end{center}
}]

\footnotetext[1]{\textit{Yu Zhang is with the Zhejiang University and with the Virtual Human Group, Netease Fuxi AI Lab. E-mail:22115031@zju.edu.cn, The work is the result of his internship at the Virtual Human Group, Netease Fuxi AI Lab.}}
\footnotetext[2]{\textit{Hao Zeng, Bowen Ma, Wei Zhang, Zhimeng Zhang and Yu Ding, Tangjie Lv, Changjie Fan are with the Virtual Human Group, Netease Fuxi AI Lab. E-mail:zenghao1110@gmail.com, \{mabowen01, zhangwei05, zhangzhimeng, dingyu01\}@corp.netease.com}}
\footnotetext[3]{\textit{$^\ast$Yu Ding is the corresponding author.}}

\markboth{Journal of \LaTeX\ Class Files,~Vol.~18, No.~9, September~2020}%
{How to Use the IEEEtran \LaTeX \ Templates}

\begin{abstract}
This work proposes a novel face-swapping framework FlowFace++, utilizing explicit semantic flow supervision and end-to-end architecture to facilitate shape-aware face-swapping. Specifically, our work pretrains a facial shape discriminator to supervise the face swapping network. The discriminator is shape-aware and relies on a semantic flow-guided operation to explicitly calculate the shape discrepancies between the target and source faces, thus optimizing the face swapping network to generate highly realistic results. The face swapping network is a stack of a pre-trained face-masked autoencoder (MAE), a cross-attention fusion module, and a convolutional decoder. The MAE provides a fine-grained facial image representation space, which is unified for the target and source faces and thus facilitates final realistic results. The cross-attention fusion module carries out the source-to-target face swapping in a fine-grained latent space while preserving other attributes of the target image (e.g. expression, head pose, hair, background, illumination, etc). Lastly, the convolutional decoder further synthesizes the swapping results according to the face-swapping latent embedding from the cross-attention fusion module. Extensive quantitative and qualitative experiments on in-the-wild faces demonstrate that our FlowFace++ outperforms the state-of-the-art significantly, particularly while the source face is obstructed by uneven lighting or angle offset.

%The proposed method effectively preserves the shape of the source faces and generates highly realistic results. 
%More specifically, our approach entails the design and training of a facial shape discriminator in the first place, which serves to explicitly model the shape discrepancy between two given input faces. Subsequently, the shape-aware flow generated by the aforementioned discriminator is integrated into the training process of the face swapping network as a form of supervision. This enables the face swapping network to generate inner-facial features that not only reflect the identity, but also the shape of the source faces. 

%We employ a pre-trained face masked autoencoder (MAE) to extract facial features from both the source face and the target face. In contrast to previous methods that use identity embedding to preserve identity information, the features extracted by our encoder can better capture facial appearances and identity information. Then, we develop a cross-attention fusion module to adaptively fuse inner facial features from the source face with the target facial attributes, thus leading to better identity preservation. 

\end{abstract}


\section{Introduction}
\IEEEPARstart{F}{ace} swapping transfers identity information from a source face onto a target face while preserving the target attributes, such as expression, pose, hair, lighting, and background. This technique has great research value due to its diverse applications in portrait reenactment, film production, and virtual reality~\cite{xu2022styleswap}. Figure \ref{fig:headline} shows several examples of face swapping.

Recent works~\cite{li2019faceshifter,chen2020simswap,xu2021facecontroller,li2021faceinpainter, zeng2022flowface} have made great improvements in achieving promising face-swapping results. However, many of them focus on transferring inner facial features, while neglecting facial contour reshaping. We are aware that facial contours also play a crucial role in conveying a person's identity, yet few efforts~\cite{wang2021hififace, zeng2022flowface} have been devoted to exploring contour transferring. Reshaping facial contours presents significant challenges as it involves making substantial changes to the pixel values. HifiFace~\cite{wang2021hififace} directly utilizes a 3D facial reconstruction model with the coarse perception of face shape. FlowFace~\cite{zeng2022flowface} adopts a two-stage framework with a specific face-reshaping network and a face-swapping network. It suffers from error accumulation introduced by the two individual stages. 
In fact, facial shape transferring is still a challenge for authentic face swapping. To further solve the shape transferring problem, we propose an end-to-end framework with the supervision of explicit semantic flow of face contour, dubbed FlowFace++. Unlike existing methods, FlowFace++ is a shape-aware and end-to-end face-swapping network.

Our end-to-end FlowFace++ is composed of three modules.
Firstly, a pre-trained masked autoencoder (MAE) is used to transform facial images into a fine-grained representation space shared by the target and source faces, which facilitates the realism of face swapping.
Then, a cross-attention fusion module performs fine-grained face swapping in latent space, including the source-to-target identity transferring, the preservation of target facial expression and other attributes (e.g., head pose, illumination, background, etc). Based on the above face-swapping latent features, a convolutional decoder is trained to carry out the synthesis of the swapped facial image. Particularly, to improve the accuracy of face contour shaping, we have developed a novel shape-aware discriminator. This discriminator relies on a semantic flow-guided operation to explicitly calculate the shape discrepancies between the target and source faces, ensuring that the face-swapping network produces an accurate face shape consistent with the given source face. Overall, our end-to-end face-swapping network achieves highly realistic and accurate face-swapping results, which benefits from the MAE encoder, cross-attention fusion module, convolutional decoder, and shape-aware discriminator. 

Prior studies\cite{wang2021hififace,chen2020simswap,kim2022smooth,li2021faceinpainter,li2019faceshifter,xu2021facecontroller} commonly utilize a face recognition model to obtain identity embedding of source face. However, given that identity embeddings are typically trained under face recognition tasks, they may not fully align with the requirements of face-swapping tasks, leading to overlooking intra-class variations~\cite{kim2022smooth}.
% In contrast, our MAE encoder is pre-trained on a large-scale face dataset~\cite{karras2017progressive,mollahosseini2017affectnet,yi2014casia,IMDB-WIKI} with a masked training strategy.
In contrast, our MAE encoder is pre-trained using a mask-then-reconstruct training strategy on a large-scale facial dataset.
%%% MBW: modified
It is capable of utilizing semantic-level features that maintain a higher degree of fine-grained information~\cite{he2022mae} than those identity-specific features of commonly-used identity embeddings.

%While other methods~\cite{yu2018bisenet} simply concatenate the identity and attribute vectors
Most previous works~\cite{li2021faceinpainter,li2019faceshifter,wang2021hififace,xu2021facecontroller} are inspired by the style transferring method to mapping target faces to the styleGAN2 latent space~\cite{karras2020analyzing} and employ AdaIN~\cite{liu2017adaptive} to integrate the identity embedding of a source face into the target face. With AdaIN, the identity embedding of a source face is viewed as image global information to adapt the space of trainable parameters, while then manipulating the latent space of the target image for the local region of face identity. Performing global operations on two distinct latent spaces does not sufficiently capture the critical interaction of local face region features in latent space. Differently, our FlowFace++ makes use of the MAE encoder to encode the source and target faces into a unified latent representation. It introduces the cross-attention fusion module to self-adaptive transfer identity information from source patches to their corresponding target patches. 
%%% MBW: representation latent space => latent representation

%\textcolor{blue}{Previous works\cite{wang2021hififace,chen2020simswap,kim2022smooth} typically rely on a face recognition model to extract the identity embedding of the source face, and then transfer it to the target. We argue that this would lose some personalized appearances during transferring, due to the identity embedding is often trained under discriminative tasks and may ignore intra-class variations~\cite{kim2022smooth}. Our face swapping network employs a pre-trained maksed autoencoder as a unified feature extractor to capture facial appearances and identity information, thus provides the possibility of preserving more fine-gained facial features during the fusion stage. Moreover, by utilizing attention mechanism~\cite{dosovitskiy2020vit} to compute patch correlations (such as nose to nose), the cross-attention fusion module of our face swapping network can adaptively combine the identity feature and attribute feature from source face and target face respectively. Comparison with previous works \cite{wang2021hififace,chen2020simswap,li2021faceinpainter,li2019faceshifter,xu2021facecontroller} that rely on AdaIN~\cite{liu2017adaptive} to inject source identity into target face, our cross-attention fusion is more flexible.}

Furthermore, our facial shape discriminator calculates pixel-level differences in facial shape between source and target faces by modeling the dense motion of facial contour. Compared to previous methods that either ignore facial shape transfer or use shape coefficients of a 3D face reconstruction model~\cite{wang2021hififace} as supervision, our facial shape discriminator provides a more finely tuned perception of facial discrepancies and facilitates the facial shape transfer capabilities of our face swapping network. 

We conduct extensive quantitative and qualitative experiments to evaluate the effectiveness of our FlowFace++ approach on in-the-wild faces. The results show that FlowFace++ outperforms the current state-of-the-art methods in terms of both objective metrics and subjective visual quality. Overall, our contributions are summerized as follows:
% \vspace{-0.03in}
\begin{itemize}
    \item We propose an end-to-end framework for shape-aware face swapping, namely FlowFace++. It can effectively transfer both the inner facial details and the facial outline of the source face to the target one, thus achieving authentic face-swapping results and robustness even under extreme input conditions (e.g. angle jamming and uneven light exposure).
    \item We design a facial shape discriminator that explicitly distinguishes the facial outline discrepancies between the given source and target input faces, with generating a shape-aware semantic flow. Our experimental results conclusively demonstrate that the incorporation of the discriminator supervision within the face swap network enables accurate facial shape transfer.
    \item We propose a pre-trained face-masked autoencoder-based face-swapping encoder (named MAE encoder), as well as a cross-attention fusion module. The MAE encoder provides a unified latent representation for the source and target face inputs. 
    %It captures not only inner facial features but also facial contour information, enabling the transfer of more comprehensive and nuanced information from the source face to the target. This approach allows for plausibility in both inner face details and facial shape, robustness even under extreme input conditions (e.g. angle jamming and uneven light exposure), enabling lifelike and convincing face swapping.
   
    %%% MBW suggestion:
    % from "face-masked autoencoder-based face-swapping encoder"
    % to "MAE-Face-based face-swapping encoder"
    % Seems cleaner, also aligns with the name used in our MAE-Face paper
\end{itemize}

% \input{02_Relatedwork}
%%%%%%%%%%%%%%%%%02_Relatedwork%%%%%%%%%%%%%%%%%%%%
\section{Related Work}
 
% The previous face swapping methods can be classified as the target attribute-guided and source identity-guided methods.
Previous face-swapping methods can be categorized as either target attribute-guided or source identity-guided approaches.

\textbf{Target attribute-guided methods} involve editing the source face first and then blending it into the target background. 
Early methods~\cite{bitouk2008face,chen2019face,lin2012face} directly warp the source face according to the target facial landmarks, thus failing to address large posture differences and expression differences. 
3DMM-based methods~\cite{blanz2004exchanging,thies2016face2face,faceswap,nirkin2018face} swap faces by 3D-fitting and re-rendering. However, these methods often struggle to handle skin color and lighting differences, leading to poor fidelity in the final result. Later, GAN-based methods improve the fidelity of the generated faces. Deepfakes~\cite{deepfakes} transfers the target attributes to the source face by an encoder-decoder structure while being constrained by two specific identities.
FSGAN\cite{nirkin2019fsgan} utilizes the target facial landmarks to animate the source face and introduces a blending network to fuse the generated source face with the target background. However, it struggles to handle significant differences in skin color. 
AOT~\cite{aot2020neurips} later concentrates on face swapping, which involves significant variations in skin color and lighting conditions by formulating appearance mapping as an optimal transport problem.
% Later, AOT~\cite{aot2020neurips} focuses on swapping faces with significant differences in skin color and lighting by formulating appearance mapping as an optimal transport problem.
Although these methods have proven effective, they still require a facial mask to blend the generated face with the target background. However, mask-guided blending can restrict the degree of face shape change, which will limit the overall quality of the final result.
% These methods always need a facial mask to blend the generated face with the target background. However, the mask-guided blending restricts the face shape change.

\textbf{Source identity-guided methods} typically rely on the use of identity embeddings or the latent representations of StyleGAN2~\cite{karras2020analyzing} to represent the source identity. These representations are then injected into the target face to transfer the source identity onto the target.
% usually adopt the identity embedding or the latent representation of StyleGAN2~\cite{karras2020analyzing} to represent the source identity and inject in into the target face. 
% FaceShifter~\cite{li2019faceshifter} designs an adaptive attentional denormalization generator to integrate the source identity embedding and the target features.
The FaceShifter model~\cite{li2019faceshifter} incorporates an adaptive attentional denormalization generator that integrates the source identity embedding and the target features to produce highly realistic facial images.
SimSwap~\cite{chen2020simswap} introduces a weak feature matching loss to effectively retain the target attributes.  
MegaFS~\cite{zhu2021one}, RAFSwap~\cite{xu2022region} and HighRes~\cite{xu2022high} leverage pre-trained StyleGAN2 models to facilitate face swapping and can achieve high-resolution face swapping. FaceController~\cite{xu2021facecontroller} exploits the identity embedding with 3D priors to represent the source identity and design a unified framework for identity swapping and attribute editing. InfoSwap~\cite{gao2021information} applies the information bottleneck principle to effectively disentangle the identity-related and identity-irrelevant information. FaceInpainter~\cite{li2021faceinpainter} also utilizes the identity embedding with 3D priors to implement controllable face in-painting under heterogeneous domains. Smooth-Swap~\cite{kim2022smooth} introduces a novel approach to constructing smooth identity embeddings, which significantly improves the training efficiency and stability of face swapping model.

However, Most existing face swapping methods do not take into account the facial outlines during face swapping. Recently, HifiFace~\cite{wang2021hififace} attempts to address this issue by introducing a 3D shape-aware identity that can control the face shape during the swapping process. However, it injects the shape representation into the latent feature space, making it difficult for the model to correctly decode the face shape.
Moreover, these methods always need a pre-trained face recognition model to extract features of source faces and another encoder for target faces during the inference time, which is not friendly to deployment.

%%%%%%%%%%%%%%%%%03_Method%%%%%%%%%%%%%%%%%%%%\input{03_Method}
\section{Proposed Method}

% \begin{figure*}[t]
% \centering
% \includegraphics[width=1.0\textwidth]{overview_v9.pdf} 
% \caption{Overview of our End-to-End FlowFace++ Framework. Firstly, we design a facial shape discriminator ($D^{shape}$), which explicitly calculates the discrepancies in facial shapes between two input faces. $D^{shape}$ generates an estimated semantic flow, representing the pixel-level motion trend from the contour of $I_2$ to that of $I_1$. Subsequently, the semantic flow $v_{so}$ between $I_s$ and $I_o$ generated by $D^{shape}$, as one of the supervisions, is incorporated into the training of the face swapping network ($F^{swa}$). $F^{swa}$ generates the inner facial details and transfers the source facial shapes by manipulating the latent face representation $e_s$ and $e_t$ using our designed cross-attention fusion module. It should be noted that \textcircled{c} in the figure represents the concatenation operation.}
% \label{fig:arch}
% \end{figure*}

\begin{figure}[t]
\setlength{\abovecaptionskip}{0.16cm}
\centering
\includegraphics[width=0.48\textwidth]{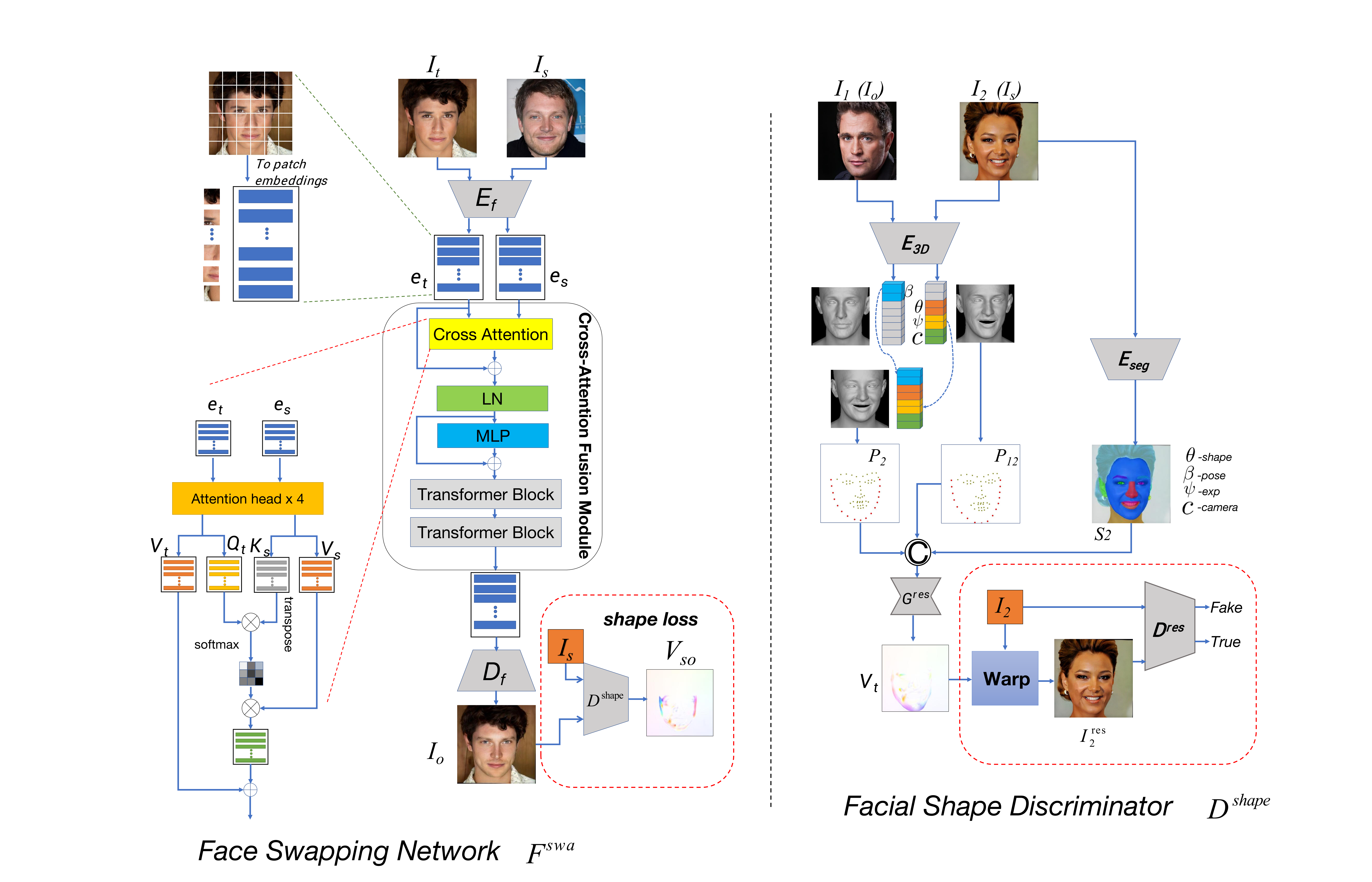} 
\caption{Overview of face swapping network ($F^{swa}$). 
$F^{swa}$ is a stack of a pre-trained face-masked autoencoder ($E^{f}$), a cross-attention fusion module, and a convolutional decoder ($D^{f}$). $F^{swa}$ generates the inner facial details and transfers the source facial shapes by manipulating the latent facial representation $e_s$ and $e_t$ using our designed cross-attention fusion module (CAFM). The training of $F^{swa}$ is supervised by a facial shape loss (Please refer to Section \ref{facial_shape_discriminator} and Figure \ref{fig:arch_Dshape} for more details.) that relies on the semantic flow $v_{so}$ of face shape from $I_o$ to $I_s$.}
\label{fig:arch_Fswa}
\end{figure}

Face swapping task aims to generate a facial image with the identity of the source face and the attributes of the target face. As shown in Figures~\ref{fig:arch_Fswa} and \ref{fig:arch_Dshape}, this paper proposes a novel face-swapping framework, named FlowFace++, to realize shape-aware face swapping. FlowFace++ consists in one face swapping network $F^{swa}$ and one facial shape discriminator $D^{shape}$. In training, $F^{swa}$ is fed with one face with target attributes and the other one with source identity. $D^{shape}$ is responsible for supervising $F^{swa}$ to perform shape-aware face swapping. 
% as well as other face identity and expression constraints. 
Once trained, $F^{swa}$ is able to carry out face swapping from one source image to another target one.

%in different identities are input to compute explicit semantic flow of facial shape $V_{t}$. 

% Let $I_{1}$ and $I_{2}$ be the first input face and the second input face of $D^{shape}$.  which utilizes explicit semantic flow supervision and end-to-end architecture   
% $D^{shape}$ explicitly models the shape discrepancy between $I_{1}$ and $I_{2}$. When training $D^{shape}$, the semantic flow generated by it explicitly warps $I_{1}$, making $I_{1}$'s facial shape converge to $I_{2}$. 
% Let $I_{s}$ be the source face and $I_{t}$ be the target face. Then $F^{swa}$ trained with supervision of the shape-aware semantic flow generated by $D^{shape}$ and other loss functions fuses the $I_{s}$'s identity information which contains both inner facial characteristics and facial shape, and attributes(e.g. expression, posture, lighting etc.) of $I_{t}$ into the result face $I_{o}$. 

% As shown in Figure~\ref{fig:arch}, FlowFace consists of a face reshaping network $F^{res}$ and a face swapping network $F^{swa}$. Let $I_{s}$ be the source face and $I_{t}$ be the target face. $F^{res}$ first transfers the shape of $I_s$ to the target face $I_t$. The reshaped image is denoted as $I^{res}_{t}$. Then $F^{swa}$ generates the inner face of $I^{res}_{t}$ and outputs the result image $I_{o}$.

\subsection{\textbf{Face Swapping Network}}
The face swapping network $F^{swa}$ is responsible for fusing identity characteristics (e.g. facial shape and inner details) of a source face and other attributes (e.g., expression, pose, hair, illumination, and others) of a target face to synthesize face-swapping results. Figure~\ref{fig:arch_Fswa} shows $F^{swa}$. Specifically, a shared MAE encoder $E_{f}$ is used to convert source face ($I_s$) and target face ($I_t$) into patch embeddings $e_s$ and $e_t$, respectively. Subsequently, a cross-attention fusion module is taken to adaptively fuse the identity information of the source face and the attribute information of the target together to produce a fusion embedding. The fusion embedding is then fed into the convolutional decoder ($D_f$), which generates the final face-swapping result $I_o$. More details will be detailed as follows.
% Then a cross-attention fusion module is designed to adaptively fuse the identity information of the source face and the attribute information of the target. Finally, the facial decoder, fed with the manipulated embeddings $e_o$, outputs the final face swapping result $I_o$.

% The face swapping network $F^{swa}$ is used to generate the inner face of $I_t^{res}$ ($I_t$). As shown in Figure~\ref{fig:arch}, we first utilize a shared face encoder $E_{f}$ to map both $I_s$ and $I_t^{res}$ into patch embeddings $e_s$ and $e_t$. Then a cross-attention fusion module is designed to adaptively fuse the identity information of the source face and the attribute information of the target. Finally, the facial decoder, fed with the manipulated embeddings $e_o$, outputs the final face swapping result $I_o$.

% \vspace{-0.1cm}
\subsubsection{Shared Face Encoder.}\label{para:encoder}
In many previous face-swapping methods, the source face is typically mapped into an ID embedding using a pre-trained face recognition model. The ID embedding is trained on recognition tasks, which leads to indistinguishable intra-class variance for a specific person. This means that the ID embedding may not fully capture the personalized appearances of a face during face swapping. This could potentially result in a loss of important details in the final face-swapping output. Additionally, previous methods use another face encoder to extract the target face attributes. This means that they propose different face representations for the source and the target faces. The distinguishable face representations bring challenges to their fusion, which may damage the fine granularity of the resulting image.

In our work, the above two issues are addressed by a pretrained face-MAE model (MAE encoder) that provides a prior fine granularity of facial representation. The pretrained face-MAE model is built with image reconstruction, instead of ID recognition. It is shared by the source and target faces, so our work provides a unified face representation for the source and target faces.
%To address the above issues, our work proposes to use a unified face representation with a face encoder shared by the source and target faces. 

%It is assumed that utilizing two different encoders is not necessary and can actually lead to increased complexity.
% We argue that utilizing two different encoders is not necessary and can actually lead to increased complexity. 
% However, we argue that using two different encoders is unnecessary and even makes deploying more complex. Moreover, the ID embedding is trained on purely discriminative tasks and may lose some personalized appearances during transferring.

Specifically, we opt to employ a shared encoder that projects both the source and target face into a unified latent representation. 
The encoder architecture is based on the MAE~\cite{he2022mae} framework and was pre-trained on a large-scale face dataset~\cite{karras2017progressive,mollahosseini2017affectnet,yi2014casia,IMDB-WIKI} consisting of 2.17 million images, utilizing a masked training strategy.
%%% MBW: similarly, might change the description according to the previous changes
When contrasted with the compact latent code of StyleGAN2~\cite{karras2020analyzing} and the identity embedding, the latent representation of MAE has the ability to more effectively capture both facial appearances and identity information. This is due to the use of masked training, which necessitates the reconstruction of masked image patches from visible neighboring patches. As a result, each patch embedding contains rich topology and semantic information. Using the pre-trained MAE encoder $E_f$, we can project a given facial image $I_*$ into a latent representation, commonly referred to as patch embeddings:
\begin{equation}
e_* = E_f(I_*),
\end{equation}
where $e_* \in R^{N*L}$. $N$ and $L$ denote the number of patches and the dimension of each embedding, respectively. 
% Therefore, we employ a shared encoder to project both the source face and the target face into a common latent representation. 
% The encoder is designed following MAE~\cite{he2022mae} and pre-trained on a large-scale face dataset using the masked training strategy.

% Compared to the compact latent code of StyleGAN2~\cite{karras2020analyzing} and the identity embedding, the latent space of MAE can better capture facial appearances and identity information, because masked training requires reconstructing masked image patches from visible neighboring patches, thus ensuring each patch embedding contains rich topology and semantic information. 

% Based on the pre-trained encoder $E_f$, we can project a facial image $I_*$ into a latent representation, also known as patch embeddings:

% The subscript $*$ is $s$ or $t$, representing the source face or the target face, respectively.

\subsubsection{Cross-Attention Fusion Module.}\label{para:cross_module}
% The obtained latent face representation $e_*$ contains complete information about a face, thus leading to the entangle of the identity and the attribute information.
% The shared face encoder projects the source face and the target face into a representative latent space. The subsequent operation is to fuse the source identity information with the target attribute in this latent space. 
Through the use of the shared MAE encoder, both the source and target faces are projected into a representational latent space. The next step is to fuse the source identity information with the target attribute within this latent space.
It's assumed that related patches, such as the nose to nose, would convey identity information during the transfer process. To account for this, we developed a cross-attention fusion module (CAFM) that can dynamically aggregate identity information from the source and blend it into the target in an adaptive and patch-wise manner.
% Intuitively, identity information should be transferred between related patches (\emph{e.g.}, nose to nose, etc.). Therefore, we design a cross-attention fusion module (CAFM) to adaptively aggregate identity information from the source and fuse it into the target.

As shown in Figure~\ref{fig:arch_Fswa}, CAFM comprises a cross-attention block and two standard transformer blocks~\cite{dosovitskiy2020vit}.
To begin, we calculate the $Q,K,V$ values for each patch embedding in the source ($e_{s}$) and target ($e_{t}$) sets. Then the cross attention is computed by:
\begin{equation}
\operatorname{CA}(Q_t, K_s)=\operatorname{softmax}\left(\frac{Q_t K_s^{T}}{\sqrt{d_{k}}}\right),
\end{equation}
In this equation, $\operatorname{CA}$ denotes Cross Attention, $Q_*, K_*, V_*$ are predicted using attention heads, and $d_k$ denotes the dimension of $K_*$.
 % where $\operatorname{CA}$ represents Cross Attention, $Q_*, K_*, V_*$ are predicted by attention heads, and $d_k$ is the dimension of $K_*$. 
The cross-attention mechanism characterizes the relationship between each target patch and the source patches. 
Subsequently, we aggregate the source identity information using the computed $\operatorname{CA}$ and fuse it with the target values through addition:
\begin{equation}
V_{fu}=\operatorname{CA} * V_s + V_t.
\end{equation}
% Then, $V_{fu}$ are normalized by a layer normalization (LN) and processed by multi-layer perceptrons (MLP). 
After that, $V_{fu}$ are normalized through a layer normalization (LN) and processed by multi-layer perceptrons (MLP).
Both the Cross Attention and MLP are accompanied by skip connections. The fused embeddings $e_{fu}$ are then fed into two transformer blocks, resulting in the final output $e_o$.
% The Cross Attention and MLP are along with skip connections. The fused embeddings $e_{fu}$ are further fed into two transformer blocks to obtain the final output $e_o$.

Finally, a convolutional decoder is utilized to generate the final swapped face image $I_o$ from the output $e_o$. In contrast to the ViT decoder in MAE, we find that the convolutional decoder produces more realistic results.
% Finally, we utilize a a convolutional decoder to generate the final swapped face image $I_o$ from $e_o$. In contrast to the ViT decoder in MAE, we find the convolutional decoder achieves more realistic results.

% \subsubsection{Facial Shape Discirminator.}\label{para:facial shape discriminator}
% Differing from the previous methods that neglect facial shape changes or use facial shape coefficients generated by 3D face reconstruction models to represent face shape changes in a less precise manner, the $D^{shape}$ has the ability to finely perceive the differences in facial shape between two input faces, as the semantic flow generated by it explicitly represents the facial outline motion trend between the two input faces. Unlike the process of training the $D^{shape}$, during the training of the $F^{swa}$, we do not require the warp process and $D^{res}$ of the $D^{shape}$. Rather, we only need to obtain the semantic flow between the source faces and resulting faces to represent the facial shape differences between them. The semantic flow is computed by:
% \begin{equation}
% % e_* = E_f(I_*),
% V_{so} = D^{shape}(I_s, I_o),
% \end{equation}

\subsubsection{Training Loss.}
To train our face-swapping network $F^{swa}$, two human face images (i.e. $I_s$ and $I_t$) will be used as the source face and target face, respectively, serving as the two inputs of $F^{swa}$. Generally, there is no ground truth available for the results of face swapping. To further constrain the output distribution of the swapping result $I_o$, the training data include a portion $25\%$ of $\{$($I_s$, $I_t$)$\}$ where $I_s=I_t$. This portion of data allows using pixel-wise reconstruction loss, as done in \cite{wang2021hififace,nirkin2019fsgan,chen2020simswap,xu2022region,li2019faceshifter,}. The other $75\%$ of the training data consists of $\{$($I_s$, $I_t$)$\}$ where $I_s$$\neq$$I_t$. 
 We design seven loss functions from the aspects of facial shape, posture, texture, expression, and identity, to constrain the result faces generated by our $F^{swa}$:
% Then we utilize six loss functions:
% To train our face swapping network ($F^{swa}$), we utilize six loss functions:
% including adversarial, reconstruction, identity, landmark, attribute, and expression loss. 
% The full objective for $F^{swa}$ can be summarised as:
\begin{equation}
\begin{split}
\mathcal{L}^{swa} =  \mathcal{L}_{adv} + \lambda_{rec}\mathcal{L}_{rec} + \lambda_{id}\mathcal{L}_{id}  +  \lambda_{exp}\mathcal{L}_{exp} \\+ \lambda_{ldmk}\mathcal{L}_{ldmk} + \lambda_{perc}\mathcal{L}_{perc} \\+ \lambda_{flow}\mathcal{L}_{flow}
\end{split}, 
\label{swap_loss_full}
\end{equation}
where $\lambda_{rec}$, $\lambda_{id}$, $\lambda_{exp}$, $\lambda_{ldmk}$, $\lambda_{attr}$ are hyperparameters for each term. We set $\lambda_{rec}$=10, $\lambda_{id}$=5, $\lambda_{exp}$=10, $\lambda_{ldmk}$=5000, $\lambda_{attr}$=2 and $\lambda_{flow}$=3 in our experiment.

% As in the face reshaping stage, the adversarial loss is used to make the resultant images more realistic, and the reconstruction loss between $I_o$ and $I_t^{res}$ is used for self-supervision since there is also no ground-truth for face swapping results. 

% During the training of $D^{shape}$, the adversarial loss is used to make the resultant images more realistic which in turn guides the generation of a more accurate semantic flow. Similarly, in the training of $F^{swa}$, we also introduce adversarial loss to generate more realistic face swapping results. 
\textit{Adversarial Loss.}
To enhance the realism of the face swapping results, we employ the hinge version adversarial loss\cite{lim2017geometric} for training, denoted by $L_{adv}$:
\begin{equation}
\mathcal{L}_{adv}=-\mathbb{E}[D^{swa}(I_o)],
\end{equation}
where $D^{swa}$ is the discriminator which is trained with:
\begin{equation}
\begin{split}
\mathcal{L}_{D}=\mathbb{E}[\max (0,1-D(I_o))]+\mathbb{E}[\max (0,1+D(I_t))] .
\end{split}
\end{equation}

% As in the face reshaping stage, the adversarial loss is used to make the resultant images more realistic, and the reconstruction loss between $I_o$ and $I_t^{res}$ is used for self-supervision since there is also no ground-truth for face swapping results. 

\textit{Reconstruction Loss.} 
% Given the absence of a ground-truth for face swapping results, 
As mentioned above, in 25\% of the training data, $I_s$ and $I_t$ are identical to each other.
%We enforce $I_s$ = $I_t$ at a rate of 25\%. 
% Since there is no ground-truth for face swapping results. We enforce $I_s$ = $I_t$ with a certain probability. 
It means that $I_t$ is the desired result of $I_o$. For these data, $I_o$ is supervised by an additional pixel-wise reconstruction loss: %and the face swapping task becomes a reconstruction task, hence we introduce a pixel-wise reconstruction loss:
\begin{equation}
\mathcal{L}_{rec}=\left\|{I_o - I_t}\right\|_2 \label{loss_rec},
\end{equation}
It is noteworthy that the reconstruction loss is not existed when $I_s$$\neq$$I_t$. 
% $I_s$ is not equal to $I_t$. 

\textit{Posture Loss.} To maintain proper face posture during face swapping, we utilize the landmark loss as a constraint:
\begin{equation}
\mathcal{L}_{ldmk}=\left\|P_{t} - P_{o}\right\|_{2},\
\end{equation}
where $P_{o}$ represents the landmarks of $I_o$. It is worth noting that only the 51 landmarks of the inner-face are included in $P_{t}$ and $P_{o}$, and the facial shape is determined by the facial shape loss (see below) rather than the landmarks of the facial contour.
% It is worth noting that the 17 landmarks on the facial contour are not included in $P_{t}$ and $P_{o}$, as their inclusion would hinder the effectiveness of facial shape loss.

\textit{Perceptual Loss.} As high-level feature maps contain semantic information, we utilize the feature maps from the final two convolutional layers of a pre-trained VGG as the representation of facial attributes. 
% Since high-level feature maps contain semantic information, we employ the feature maps from the last two convolutional layers of pre-trained VGG as the facial attribute representation. 
The loss is formulated as:
\begin{equation}
\mathcal{L}_{perc}=\left\|VGG(I_t) - VGG(I_o)\right\|_{2}.\
\end{equation}

\textit{Expression Loss.} 
% To make the expression of the swapped face $I_o$ more consistent with the target face, 
We adopt a novel fine-grained expression loss~\cite{zhang2021learning} that penalizes the $\mathcal{L}_2$ distance of two expression embeddings:
\begin{equation}
\mathcal{L}_{exp} = \left\|{E_{exp}(I_o) - E_{exp}(I_t)}\right\|_2. \label{loss_exp}
\end{equation}

\textit{Identity Loss.} The identity loss is utilized to enhance the identity similarity between $I_s$ and $I_o$:
\begin{equation}
\mathcal{L}_{id} = 1- cos(E_{id}(I_o), E_{id}(I_s)), \label{loss_id}
\end{equation}
where $E_{id}$ denotes a face recognition model~\cite{deng2019arcface} and $cos$ denotes the cosine similarity. 

\textit{Facial Shape Loss.} The facial shape loss is used to constrain $I_o$ to have a similar facial shape as $I_s$. 
% The shape-aware semantic flow explicitly calculates the discrepancies in facial shape between the two input faces, thereby quantifying the shape differences between them.
The shape-aware semantic flow generated by $D_{shape}$ explicitly quantifies the discrepancies in facial contour between the two input faces by modeling the pixel-level motion trend on facial shape. 
Consequently, as the face shape of $I_{o}$ approaches that of $I_{s}$, the semantic flow between $I_{s}$ and $I_{o}$ tends towards zero:
\begin{equation}
V_{so} = D^{shape}(I_s, I_o),\
\end{equation}
\begin{equation}
\mathcal{L}_{flow} = \left\|V_{so} - zeros\_like(V_{so})\right\|_{2},\
\end{equation}
where $V_{so}$ denotes the semantic flow between $I_{s}$ and $I_{o}$, $zeros\_like(V_{so})$ denotes a full-zero matrix  with the same dimensions as $V_{so}$ and $\left\|*\right\|_2$ denotes the euclidean distance. 
% where $E_{id}$ denotes a face recognition model~\cite{deng2019arcface} and $cos$ denotes the cosine similarity. 

\subsection{\textbf{facial shape discriminator}}
\label{facial_shape_discriminator}
%%% high level
As mentioned above, a facial shape discriminator, $D^{shape}$, is built to explicitly evaluate the face shape contour. It is used to constrain the resulting face to have a face contour close to the source face as much as possible. When training $F^{swa}$, $D^{shape}$ quantifies the discrepancy of the facial contours of the resulting face and the source face. The quantified discrepancy is viewed as the facial shape loss, $\mathcal{L}_{flow}$, which is mentioned above. Specifically, the discrepancy is based on the estimated semantic flow which reflects the pixel-level motion from the resulting contour to the source one.
% Specifically, the evaluation relies on the semantic flow of face shape contour. The building of $D^{shape}$ is based on the estimated semantic flow which reflects the pixel-level motion from the resulting contour to the source one. 
% The estimated semantic is used to explicitly warps the input face while training $F^{swa}$.

\begin{figure}[t]
\setlength{\abovecaptionskip}{0.16cm}
\centering
\includegraphics[width=\linewidth]{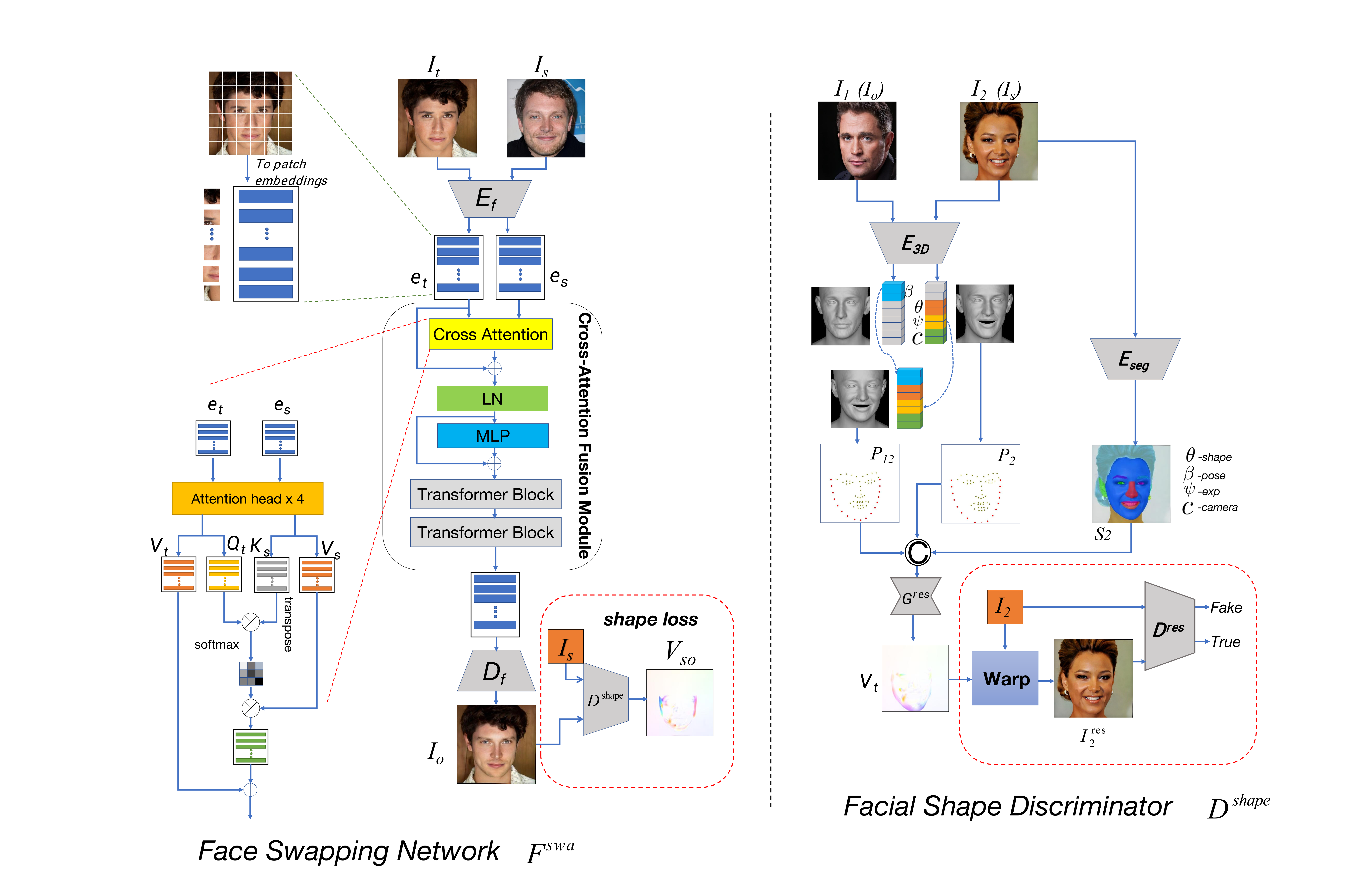} 
\caption{Overview of facial shape discriminator ($D^{shape}$). $D^{shape}$ explicitly calculates the discrepancies in facial shapes between two input faces (i.e. $I_1$ and $I_2$). It generates a semantic flow $V_t$ that represents the pixel-level motion trend from the contour of $I_2$ to that of $I_1$. During the training process of $D^{shape}$ itself, $V_t$ is used to explicitly warp $I_2$ to obtain $I_{2}^{res}$, as shown in the red box. The discrepancies between $I_2$ and $I_{2}^{res}$ form the loss functions to supervise the training of $D^{shape}$.
%in order to calculate the loss and generate more accurate semantic flow. 
Once trained, $D^{shape}$ is used to supervise the training of $F^{swa}$ by replacing $I_1$ and $I_2$ with the output face $I_o$ and the source face $I_s$ in $F^{swa}$. The calculated semantic flow ($V_{so}$) is taken as the facial shape loss for the training of $F^{swa}$, leaving out the wrapping process marked with the red box.}
\label{fig:arch_Dshape}
\end{figure}

The building of $D^{shape}$ relies on training a neural network model in a setting of pairs of two images ($I_1$ and $I_2$). The images in a pair are randomly selected from three widely-used face datasets: CelebA-HQ~\cite{karras2017progressive}, FFHQ~\cite{karras2019style}, and VGGFace2~\cite{cao2018vggface2}.

In the training of $D^{shape}$, the semantic flow of face contour shape is estimated from $I_2$ to $I_1$. Then the flow warps the shape of $I_2$ pixel-wisely to make its contour converge to the $I_1$. Thus the semantic flow reflects the pixel-level motion of facial contour and then achieves fine modeling and perception of discrepancies in facial shape. To achieve this goal, $D^{shape}$ requires a face shape representation of the two input faces and then estimates a semantic flow according to the differences of two shape representations.

In the training of $F^{swa}$, $D^{shape}$ is utilized as the facial shape loss. Specifically, $I_{o}$ and $I_{s}$ are at the place of $I_1$ and $I_2$, respectively.

% \textit{When $D^{shape}$ is training, the estimated semantic flow generated by it warps shape of the second input face ($I_2$) pixel-wisely to make it's contour converging to the first input face ($I_1$). }

%We design the facial shape discriminator, $D^{shape}$, to explicitly calculate the discrepancies between two input faces. When $D^{shape}$ is training, the estimated semantic flow generated by it warps shape of the second input face ($I_2$) pixel-wisely to make it's contour converging to the first input face ($I_1$). Thus semantic flow reflects the pixel-level motion of facial contour and then achieves fine modeling and perception of discrepancies of facial shape. To achieve this goal, $D^{shape}$ requires a face shape representation of the two input faces. Then it estimates a semantic flow according to the differences of the shape representation. 
% Finally, the semantic flow is used to supervise the $F^{swa}$ when training it. 

% We design the face reshaping network, $F^{res}$, to address the shape discrepancy between the source and target faces. It warps the target face shape explicitly pixel-wise with an estimated semantic flow. To achieve this goal, $F^{res}$ requires a face shape representation that models the shape differences between the source and target faces. Then it estimates a semantic flow according to the above shape differences. Finally, the semantic flow is used to warp the target face shape.

\subsubsection{Face Shape Representation.}\label{para:face_shape_representation} 
Since our facial shape discriminator needs to warp the face shape pixel-wisely while being trained, we choose the explicit facial landmarks as the shape representation. We use a 3D face reconstruction model (3DMM)~\cite{DECA:Siggraph2021} to obtain facial landmarks. As shown in Figure~\ref{fig:arch_Dshape}, the 3D face reconstruction model $E_{3D}$ extracts 3D coefficients of the source and target: 
% faces, respectively:
\begin{equation}
\begin{split}
(\beta_*, \theta_*, \psi_*, c_*)=E_{3D}(I_*),
\end{split}
\end{equation}
where $\beta_*, \theta_*, \psi_*, c_*$ are the FLAME coefficients~\cite{FLAME:SiggraphAsia2017} representing the facial shape, pose, expression, and camera, respectively. $*$ is $1$ or $2$, representing the first or the second input face. With these coefficients, the second input face can be modeled as:
\begin{equation}
% M_t(\beta_t, \theta_t, \psi_t)=W\left(T_P(\beta_t, \theta_t, \psi_t), \mathbf{J}(\beta_t), \theta_t, \mathcal{W}\right),
M_2(\beta_2, \theta_2, \psi_2)=W\left(T_P(\beta_2, \theta_2, \psi_2), \mathbf{J}(\beta_2), \theta_2, \mathcal{W}\right),
\end{equation}
where $M_2$ represents the 3D face mesh of the $I_2$. $W$ is a linear blend skinning (LBS) function that is applied to rotate the vertices of $T_P$ around joint $J$. $\mathcal{W}$ is the blend weights. $T_P$ denotes the template mesh $\overline{T}$ with shape, pose, and expression offsets~\cite{FLAME:SiggraphAsia2017}.
% \begin{equation}
% T(\beta, \theta, \psi)=\overline{T}+B_{sh} \beta+B_{po} \theta +B_{ex} \psi,
% \end{equation}
% where $B_{sh}$, $B_{po}$ and $B_{ex}$ are the bases of face shapes, poses and expressions. 

\begin{figure*}[t]
\centering
\includegraphics[width=0.98\textwidth]{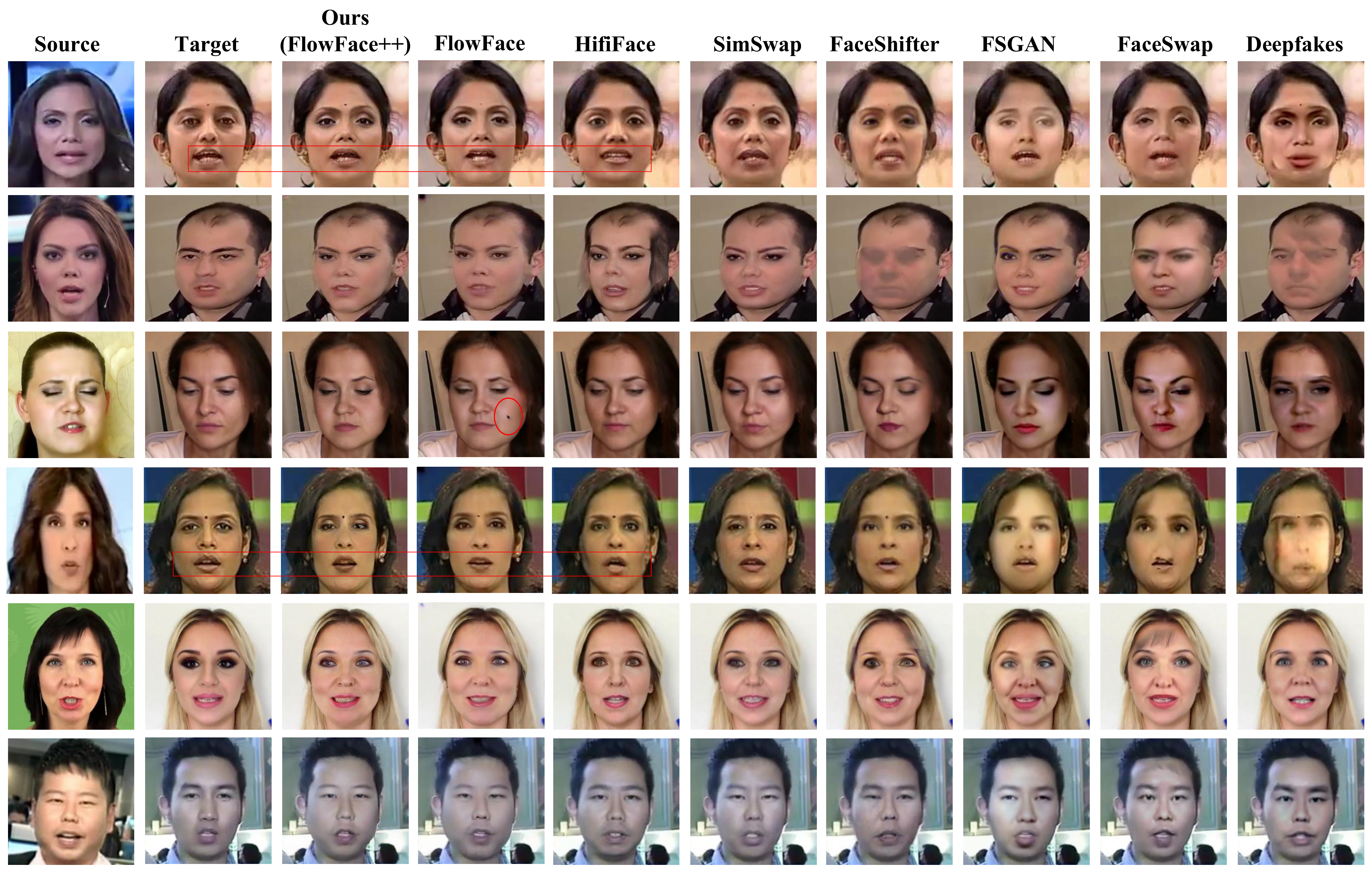}
\caption{Qualitative comparisons with Deepfakes, FaceSwap, FSGAN, FaceShifter, SimSwap (SS) and HifiFace on FF++. Our FlowFace++ outperforms the other methods significantly, especially in preserving face shapes, identities, and expressions. 
}
\label{fig:FF_comparison}
\end{figure*}

Then, we reconstruct $I_1$ similarly, except that the pose and expression coefficients are replaced with the $I_2$'s ones. The obtained 3D face mesh is denoted as $M_{12}$. Finally, we sample 3D facial landmarks from $M_2$ and $M_{12}$ and project these 3D points to 2D facial landmarks with the target camera parameter $c_2$:
\begin{equation}
\begin{split}
    P_{2} =s \Pi\left(M_{2}^{i}\right)+t, \\
    P_{12} =s \Pi\left(M_{12}^{i}\right)+t,
\label{con:landmarks}
\end{split}
\end{equation}
where $M_{*}^{i}$ is a vertex in $M_{*}$, $\Pi$ is an orthographic 3D-2D projection matrix, and $s$ and $t$ are parameters in $c_2$, indicating isotropic scale and 2D translation. $P_{*}$ denotes the 2D facial landmarks. It should be noted that we only use the landmarks at the facial contours as the shape representation since inner facial landmarks contain identity information that may influence the reshaping result.

\subsubsection{Semantic Flow Estimation.}\label{para:se_flow_estimation}
% Although facial landmarks can be directly used to compute the warping flow, the computed flow is not suitable for pixel-wise warping since landmarks are sparse representations. Therefore,
% Although we can directly use the relative displacement between $P_t$ and $P_{s2t}$ to warp the target face, the sparse displacement cannot handle the pixel-wised motion well.
The relative displacement between $P_2$ and $P_{12}$ only describes sparse movement. To accurately perceive the discrepancies of the two faces, we need to obtain dense motion between them. Therefore, we propose the semantic flow, which models the semantic correspondences between two faces. To achieve a more shape-aware semantic flow, $D_{shape}$ warps the target face according to the semantic flow during training, and constrains the warped target face to be consistent with the source face in terms of facial shape. We design a semantic guided generator $G^{res}$ to estimate the semantic flow. Specifically, $G^{res}$ requires three inputs: $P_{12}$, $P_2$ and $S_2$, where $P_{12}$ and $P_2$ are the 2D facial landmarks obtained above.
% between the source and the target face. 
$S_2$ is the second face segmentation map that complements the semantic information lost in facial landmarks. The output of $G^{res}$ is the estimated semantic flow $V_t$, the formulation is:
\begin{equation}
V_t = G^{res}(P_{12}, P_2, S_2).
\end{equation}

Then, a warping module is introduced to generate the warped faces using $V_t$. We find that an inaccurate flow is likely to produce unnatural images, On the contrary, imposing constraints on the warped images can lead to a more precise semantic flow, and therefore, we design a semantic guided discriminator $D^{res}$ that ensures $G^{res}$ to produce a more accurate flow. The warping operation is conducted on $I_2$:
\begin{equation}
I_2^{res} = F(V_t, I_2),
% (I_t^{res},S_t^{res}) = F(V_t, I_t, S_t),
% (I_t^{res}) = T(V_t, I_t)
\end{equation}
Where $F$ is the warping function in the warping module. We feed the warped face $I_2^{res}$ to $D^{res}$. Thus, $D^{res}$ is able to discriminate whether the input is real or fake. It should be noted that $D^{res}$ and warping module are only used during training of $D^{shape}$.

\subsubsection{Training Loss.}
We adopt three loss functions for $D^{shape}$: 
% reconstruction loss, landmark loss, and adversarial loss. The full objective for $F_{res}$ is denoted as:
\begin{equation}
\mathcal{L}^{res} = 
\mathcal{L}_{adv} + \lambda_{rec}\mathcal{L}_{rec} + \lambda_{ldmk}\mathcal{L}_{ldmk},
\label{res_loss_full}
\end{equation}
% where $\lambda_{ldmk}$ and $\lambda_{rec}$ are hyperparameters for each term. In our experiments, we set $\lambda_{ldmk}$=800 and $\lambda_{rec}$=10.
where $\lambda_{ldmk}$ and $\lambda_{rec}$ represent hyperparameters associated with each term. In our experimental setup, we have designated the value of $\lambda_{ldmk}$ as 800 and that of $\lambda_{rec}$ as 10.

As in the training of $F^{swa}$, the reconstruction loss between $I_2^{res}$ and $I_2$ is used for self-supervision since there is also no ground-truth for face swapping results.

\textit{Adversarial Loss.}
The more realistic the resultant images are, the more accurate the generated shape-aware semantic flows are, therefore we employ the hinge version adversarial loss\cite{lim2017geometric} for training, denoted by $L_{adv}$:
\begin{equation}
\mathcal{L}_{adv}=-\mathbb{E}[D^{res}(I_2^{res})],
\end{equation}
where $D^{res}$ is the discriminator which is trained with:
\begin{equation}
\begin{split}
\mathcal{L}_{D}=\mathbb{E}[\max (0,1-D(I_2))]\\+\mathbb{E}[\max (0,1+D(I_2^{res}))] .
\end{split}
\end{equation}

% \textit{Reconstruction Loss.}
% As in the Since there is no ground-truth for face warped results, 
% % we enforce $I_1 = I_2$ with a certain probability 
% we incorporate a mechanism to ensure that $I_1$ and $I_2$ are equivalent with a certain probability when training $G^{res}$. Then the face warping task becomes a reconstruction task, and we introduce a pixel-wise reconstruction loss:
% \begin{equation}
% \mathcal{L}_{rec}=\left\|{I_2^{res} - I_2}\right\|_2 \label{loss_rec},
% \end{equation}
% where $\left\|*\right\|_2$ denotes the euclidean distance.

\textit{Landmark Loss.} Since there is not pixel-wised ground truth for $I_2^{res}$, we rely on the 2D facial landmarks $P_{12}$ to regulate the shape of $I_2^{res}$. To be specific, we employ a pre-trained facial landmark detector~\cite{sun2019high} to forecast the facial landmarks of $I_2^{res}$, which are denoted as $P_{2}^{res}$. Then the loss is computed as:
\begin{equation}
% \mathcal{L}_{ldmk}=\frac{1}{N} \sum_{i=1}^{N}\left\|P_{pred}^{i}-P_{s2t}^{i}\right\|_{1}
\mathcal{L}_{ldmk}=\left\|P_{2}^{res} - P_{12}\right\|_{2}.\
\end{equation}
% where $P_{pred}^{i}$ is the $i_{th}$ point of $P_{pred}$ and $P_{s2t}^{i}$ is the $i_{th}$ point of $P_{s2t}$.

At this point, our designed facial shape discriminator is able to generate a shape-aware semantic flow which finely perceives discrepancies of facial shape between two input faces. Subsequently, 
the semantic flow can be utilized to enforce similarity between the facial shape of the face-swapped faces and that of the source faces during training of the $F_{swa}$.
% the semantic flow can be used to constraint the facial shape of result images is similar to that of source faces when train the $F_{swa}$. 

\section{Experiments}
% \normalsize
\begin{table}[t]
\setlength{\abovecaptionskip}{0.2cm}
\centering
% \caption{Quantitative comparisons with state-of-the-art methods on FF++. "\dag" means the results are cited from their papers.}

\resizebox{0.5\textwidth}{!}{
\begin{tabular}{c|ccc|c|c|c}
\multirow{2}{*}{Methods} & \multicolumn{3}{c|}{ID Acc($\%$) $\uparrow$}                     & \multicolumn{1}{l|}{\multirow{2}{*}{Shape$\downarrow$}} & \multicolumn{1}{l|}{\multirow{2}{*}{Expr.$\downarrow$}} & \multicolumn{1}{l}{\multirow{2}{*}{Pose.$\downarrow$}} \\ \cline{2-4}
                         & CosFace  & SphereFace & Avg   & \multicolumn{1}{l|}{}                       & \multicolumn{1}{l|}{}                       & \multicolumn{1}{l}{}                       \\ \hline
Deepfakes              & 83.32     & 86.93      & 85.13 & 1.78                             & 0.57                             & 4.05                             \\
FaceSwap               & 70.74    & 76.69      & 73.72 & 1.85                             & 0.43                             & 2.20                             \\
FSGAN                  & 48.88     & 54.09      & 51.49 & 2.18                             & 0.30                             & 2.20                             \\
FaceShifter            & 97.38\dag     & 80.64      & 89.01 & 1.68                             & 0.36                             & 2.28                             \\
SimSwap                & 93.37      & 96.15      & 94.76 & 1.74                             & 0.30                             & \textbf{1.40}  \\ \hline                        
HifiFace               & 98.48\dag     & 90.61      & 94.55 & 1.62                             & 0.33                            & 2.30 \\
FlowFace       & \underline{99.20}     & \underline{98.87}      & \underline{99.04} & \textbf{1.17}\dag                              & \underline{0.24}                             & 2.40     
\\ \hline

% $F^{swa}$          & \underline{99.18}      & \underline{98.23}      & \underline{98.70} & \underline{1.43}                             & \textbf{0.21}                             & \underline{1.99}                             \\
Ours & \textbf{99.51}      & \textbf{99.03}   & \textbf{99.27} & \underline{1.43}  & \textbf{0.23}   & \underline{2.20}             
\end{tabular}
}

\caption{Quantitative comparisons with other methods on FF++ dataset. "\dag" means the results are from their papers.}
\label{tab:quantitative_comparison}
\end{table}

To validate our FlowFace++ method, we perform quantitative and qualitative  comparisons with state-of-the-art approaches, as well as a user study. Additionally, we conduct several ablation experiments involving those employed $D^{shape}$, CAFM, MAE, and convolutional decoder to validate our design.
% Our method is validated through qualitative and quantitative comparisons with state-of-the-art ones and a user study. Moreover, several ablation experiments are also reported to validate our design of XxxFace, involving $D^{shape}$, CAFM, MAE and convolutional decoder.

\subsection{\textbf{Implementation Details}}
\noindent{\textit{Dataset.}} We collect the training dataset from three widely-used face datasets: CelebA-HQ~\cite{karras2017progressive}, FFHQ~\cite{karras2019style}, and VGGFace2~\cite{cao2018vggface2}. The faces are first aligned and cropped to $256\times256$.
% The training dataset is collected from three commonly-used face datasets: CelebA-HQ~\cite{karras2017progressive}, FFHQ~\cite{karras2019style}, and VGGFace2~\cite{cao2018vggface2}. Faces are aligned and cropped to $256\times256$.
To ensure high-quality training, we filter out low-quality images from the above datasets. The final used dataset consists of 350K high-quality face images, and we randomly select 10K images from the dataset as the validation dataset. For the comparison experiments, we construct the test set by sampling FaceForensics++(FF++)~\cite{roessler2019faceforensicspp}, following the methodology used in~\cite{li2019faceshifter}. The FF++ dataset comprises of 1000 video clips, and we collect the test set by sampling ten frames from each clip, resulting in a total of 10000 images.  

\noindent{\textit{Training.}} Our FlowFace++ is trained in two stages. Specifically, $D^{shape}$ is first trained for 250K steps with a batch size of eight. As for $F^{swa}$, we first pre-train the MAE encoder following the training strategy of MAE on our face dataset. Then we fix the MAE encoder and train other components of $F^{swa}$ for 640K steps with a batch size of eight. Due to the time-consuming nature of extracting coefficients from the 3D face reconstruction model used in $D^{shape}$, the facial shape loss is not involved in the training for the first 320K steps in order to accelerate the training speed.
% Our FlowFace is trained in a two-stage manner. Specifically, $F^{res}$ is first trained for 32K steps with a batch size of eight. As for $F^{swa}$, we first pre-trained the face encoder following the training strategy of MAE on our face dataset. Then we fix the encoder and train other components of $F^{swa}$ for 640K steps with a batch size of eight. 
We utilize the Adam optimizer~\cite{kingma2014adam}, with $\beta_1$ set to 0 and $\beta_2$ set to 0.99, and a learning rate of 0.0001.  
% More details are in the supplementary materials and our codes will be made publicly available upon publication of the paper.

\noindent{\textit{Metrics.}} We employ four metrics for the quantitative evaluation of our model: identity retrieval accuracy (ID Acc), shape error, expression error (Expr Error), and pose error.
% The shape error is used to measure the shape difference between the swapped face and the source face, and the pose error is used to measure the pose distance between the swapped face and the target face. 
We follow the same testing protocol as outlined in ~\cite{li2019faceshifter,wang2021hififace}. However, since certain pre-trained models used as metrics in their evaluation are not accessible, we utilize other models for evaluation. For ID Acc, we employ two other face recognition models: CosFace (CF)~\cite{wang2018cosface} and SphereFace (SF)~\cite{liu2017sphereface}, to perform identity retrieval for a more comprehensive comparison. For expression error, we adopt another expression embedding model~\cite{vemulapalli2019compact} to compute the euclidean distance of expression embeddings between the target and swapped faces.

\begin{figure*}[t]
\centering
\includegraphics[width=1.0\textwidth]{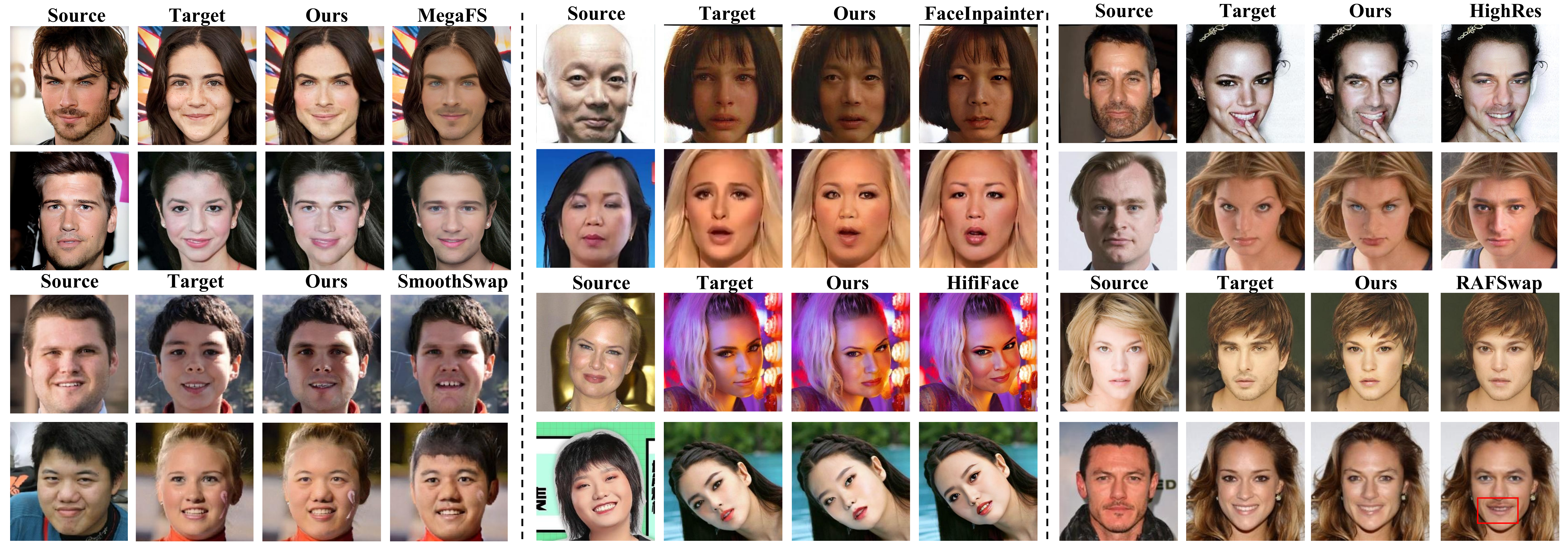}
\caption{Qualitative comparisons with more methods including  MegaFS~\cite{zhu2021one}, FaceInpainter~\cite{li2021faceinpainter}, HighRes~\cite{xu2022high} and SmoothSwap~\cite{kim2022smooth}. The images of the compared methods shown in our paper are cropped from either their original papers or from released results.
}
\label{fig:additional_qualitative}
\end{figure*}

% \begin{figure*}[t]
% \centering
% \includegraphics[width=1.0\textwidth]{temporal_v2.pdf}
% \caption{Qualitative comparisons with more methods including  FlowFace, SimSwap, FSGAN, HifiFace, FaceShifter and FaceSwap. The sampling interval was set to 10 frames. The video was sourced from short video websites and movie clips, and can be enlarged to view more details.
% }
% \label{fig:temporal_compare}
% \end{figure*}

\subsection{\textbf{Comparisons with State-of-the-arts}}

\subsubsection{Quantitative Comparisons.} 
Our method is compared with seven methods including Deepfakes~\cite{deepfakes}, FaceSwap~\cite{faceswap}, FlowFace~\cite{zeng2022flowface}, FSGAN~\cite{nirkin2019fsgan}, FaceShifter~\cite{li2019faceshifter}, SimSwap~\cite{chen2020simswap},  and HifiFace~\cite{wang2021hififace}.
For Deepfakes, FaceSwap, FaceShifter, and HifiFace, we use their released face swapping results of the sampled 10,000 images. For FlowFace, FSGAN and SimSwap, the face swapping results are generated with their released codes. 

Table~\ref{tab:quantitative_comparison} demonstrates that the proposed FlowFace++ outperforms the other methods in most evaluation metrics, including ID Acc, shape error, and expression error (Expr Error). These results validate the superiority of FlowFace++. FlowFace++ produces slightly worse results in terms of pose error compared to other methods, which may be attributed to our manipulation of face shape that poses greater challenges in pose. The employed head pose estimator is sensitive to face shapes, which might have impacted the results.
% hallenging in pose.

% \vspace{-0.2cm}
\subsubsection{Qualitative Comparisons.} The qualitative comparisons are conducted on the same FF++ test set collected in the quantitative comparisons. As shown in Figure~\ref{fig:FF_comparison}, our FlowFace++ (or FlowFace) maintains the best face shape consistency. Most methods result in face shapes similar to the target ones since they do nothing to transfer the face shape.

Although HifiFace is intentionally designed to manipulate the face shape, our method still outperforms it in terms of evaluation metrics. Compared to HifiFace, Figure~\ref{fig:FF_comparison} illustrates that our generated face shapes are more similar to the source faces. Since HifiFace injects the shape representation into the latent feature space, and directly uses facial landmarks generated by the 3D face reconstruction module as supervision for facial shape, it may be harder to achieve a fine perception of facial shape than our explicit supervision of facial shape differences by the semantic flow.
% Since HifiFace injects the shape representation into the latent feature space, it is harder to accurately decode the face shape from the latent feature than our explicit semantic flow.
Additionally, our proposed method excels at preserving fine-grained target expressions, as marked indicated by the red boxes in Rows 1 and 4 of Figure~\ref{fig:FF_comparison}. 

\begin{table}[t]
\setlength{\abovecaptionskip}{0.2cm}
% \caption{Subjective comparisons with SimSwap and HifiFace on FF++.}
\centering
\resizebox{0.5\textwidth}{!}{
\begin{tabular}{l|c|c|c|c}

Method      & Shape. ($\%$)$ \uparrow$  & ID. ($\%$)$ \uparrow$  & Exp. ($\%$)$ \uparrow$ & Realism ($\%$)$ \uparrow$ \\ \hline
SimSwap    & 21.89    & 24.41 & \textbf{44.37} & 24.68    \\ 
HifiFace & 36.44 & 33.53 & 15.98 & 36.23    \\ \hline
Ours        & \textbf{41.67} & \textbf{42.06} & 39.66 & \textbf{39.09} \\
\end{tabular}
}
\caption{Subjective comparisons with SimSwap and HifiFace on FF++.}
\label{tab:subjective_comparison}
\end{table}

We further compare our methods with more six SOTA face swapping methods:MegaFS~\cite{zhu2021one}, FaceInpainter~\cite{li2021faceinpainter}, HighRes~\cite{xu2022high}, SmoothSwap~\cite{kim2022smooth}, HifiFace~\cite{wang2021hififace} and RAFSwap~\cite{xu2022region}. 
% Due to the unavailability of the official code, 
The source, target faces and results used in this comparative experiment are cropped from the original papers of these methods. Figure~\ref{fig:additional_qualitative} demonstrates that our method is capable of better transferring the shape of the source face to the target. Although the results exhibited in the paper of HifiFace demonstrate its effectiveness in transferring facial shape, it suffers from the problem of facial expression leakage from the source face to the result. While SmoothSwap can change the facial shape, it often destroys the target attributes (\emph{e.g.}, hairstyle, and hair color). 

The qualitative comparisons above also demonstrate that our results exhibit higher similarity to the source face in terms of inner facial features (\emph{e.g.}, beard), confirming that our MAE encoder is more efficacious in effectively representing facial appearances than the identity embedding used in \cite{wang2021hififace,li2021faceinpainter,kim2022smooth} or the latent code of StyleGAN2 used in \cite{zhu2021one,xu2022high,xu2022region}. Moreover, our approach demonstrates a higher degree of fidelity in preserving target attributes (\emph{e.g.}, skin color, lighting, and expression), in comparison to other other methods.

% \vspace{-0.2cm}
% 这段需要换成我们自己的用户测试部分
\subsubsection{User Study.}
In order to further validate our FlowFace++, we conduct a subjective comparison study with two of the state-of-the-art face swapping methods, SimSwap and HifiFace, both of which have publicly shared their codes or results. We randomly select 30 instances of swapped faces generated from each of the three aforementioned methods. Participants are instructed to choose the best results in terms of shape consistency, identity consistency, expression consistency, or image realism. 

% 我们为每张图像设置最多39位参与者进行评判，在每个参与者选出答案后，对一个确定的选项，选择该项的记为1，不选择该项的记为0，在置信度阈值为80%的条件下，如果存在一个选项的置信区间左端点>0.5，则认为这张图像已经完成了评判过程。否则持续增加参与者人数，直到39个人的上限
For each image, a maximum of 39 participants are recruited for evaluation. Each participant selects their preferred option. And for a given option, assigns a value of 1 if they select it and 0 if they do not. Under the condition of a confidence threshold of 80\%, if the left endpoint of a confidence interval for one given option is greater than 0.5, this image is considered to have completed the evaluation process. Otherwise, the number of participants is increased until the maximum of 39 participants was reached. 
Table~\ref{tab:subjective_comparison} shows that our method outperforms the two baselines in terms of shape consistency, identity consistency and image realism, validating the superiority of our method. In terms of expression consistency, our FlowFace++ slightly lags behind SimSwap, which could be attributed to the coupling between facial shape and expression, where changes in facial shape may affect the discriminability of expressions.

\subsection{\textbf{Robustness Comparisons.}}
We compare the performance differences between our FlowFace++ and four other available face swap methods under extreme input conditions. It's worth noting that FaceShifter relies on an open-source implementation \footnote{https://github.com/mindslab-ai/hififace} by others, rather than authors. 

\begin{figure}[t]
\setlength{\abovecaptionskip}{0.16cm}
\centering
\includegraphics[width=0.48\textwidth]{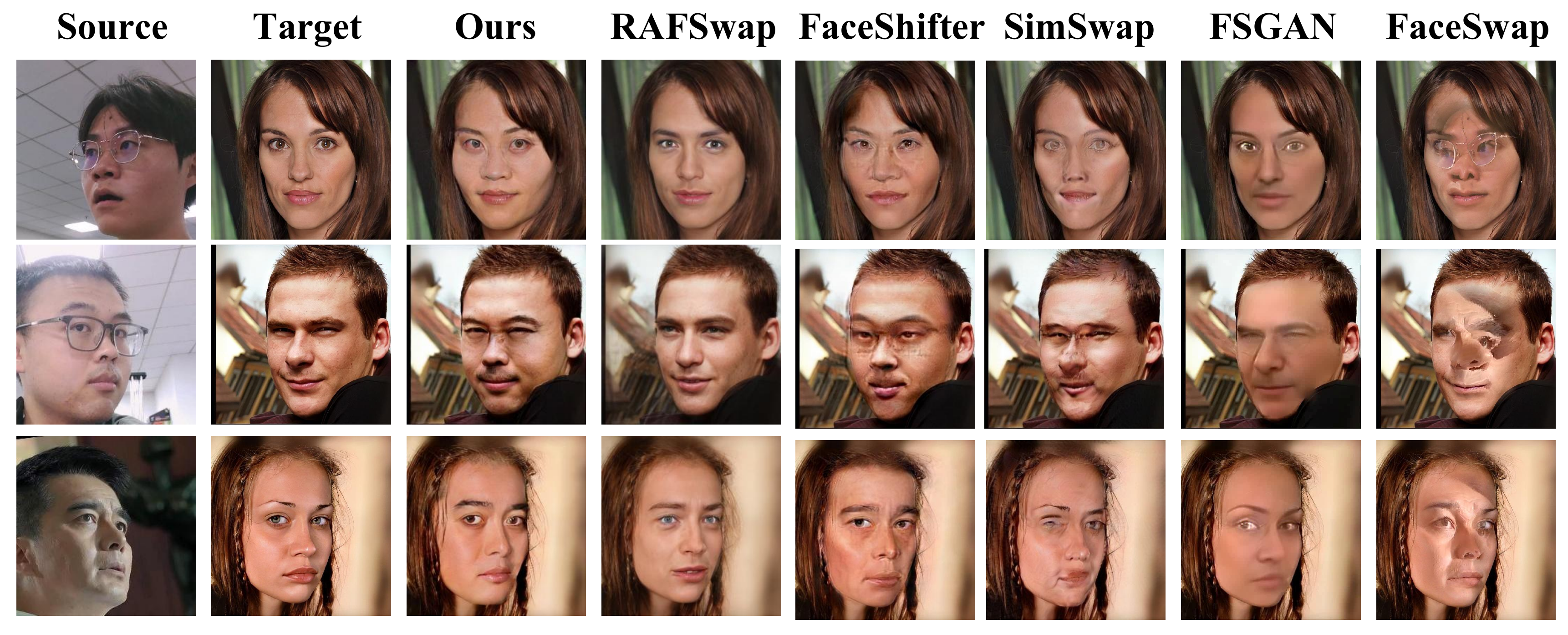}  
\caption{Qualitative comparison of our FlowFace++ with SimSwap\cite{chen2020simswap}, FSGAN\cite{nirkin2019fsgan}, FaceSwap\cite{faceswap}, and FaceShifter\cite{li2019faceshifter} under extreme input conditions, where the source faces exhibit large angular deviations. The FaceShifter utilizes an open-source implementation by others.
}
% \vspace{-1em}
\label{fig:angle_compare}
\end{figure}

\subsubsection{Angle jamming.}As shown in Figure~\ref{fig:angle_compare},  FSGAN and FaceSwap struggle to produce convincing results from the large angular deviations of the source faces. On the other hand, FaceShifter and SimSwap generate resulting faces that are not as clear and accurate. Although other results generated by RAFSwap based on StyleGAN2 are still clear, it cannot effectively extract the identity information of source faces, resulting in poor identity similarity between the result and source faces. In contrast, our FlowFace++ is still able to transfer the attributes and facial shape of the source faces to the target faces accurately. 

\begin{figure}[t]
\setlength{\abovecaptionskip}{0.16cm}
\centering
\includegraphics[width=0.48\textwidth]{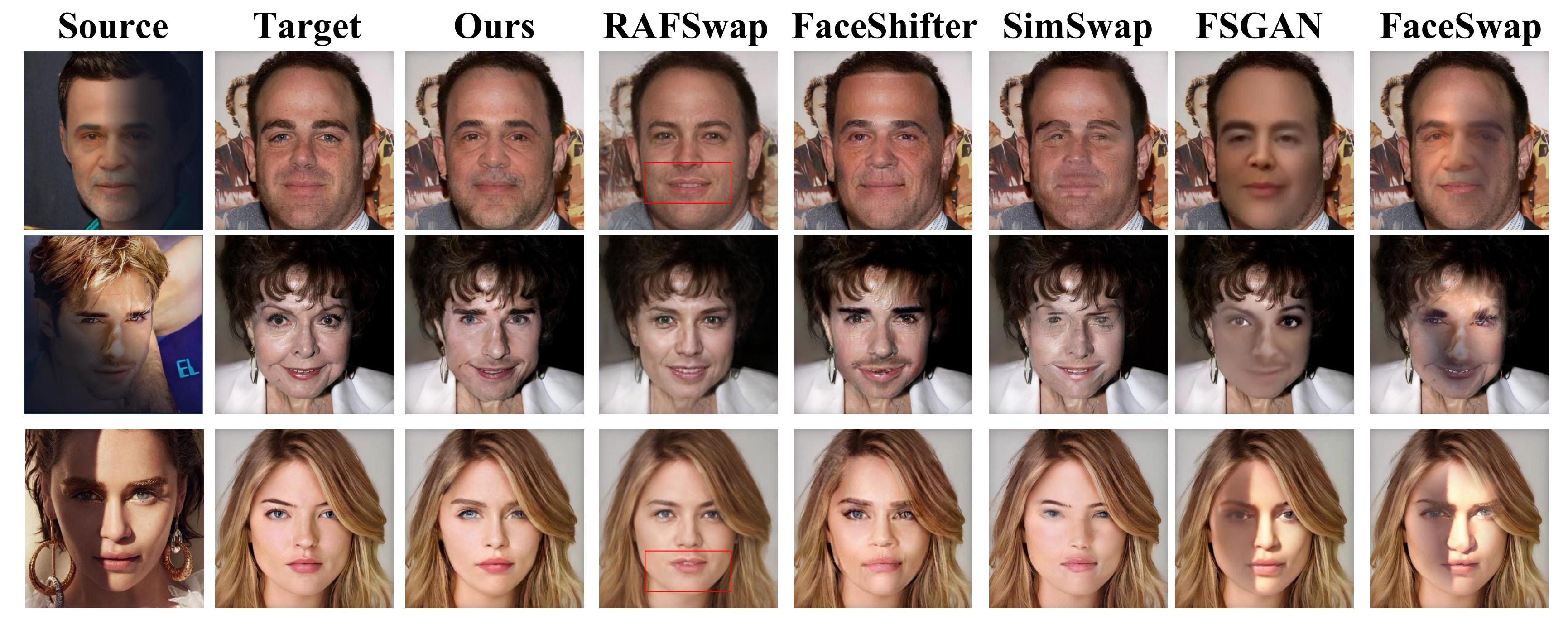}  
\caption{Qualitative comparison of our FlowFace++ with SimSwap, FSGAN, FaceSwap, and FaceShifter under conditions of uneven illumination in the source faces. 
% The FaceShifter utilizes a unofficial open-source implementation.
}
% \vspace{-1em}
\label{fig:shadow_compare}
\end{figure}

\subsubsection{Uneven light exposure. }As shown in Figure~\ref{fig:shadow_compare}, FSGAN and FaceSwap mistakenly transfer the lighting of the source faces to the result faces, while SimSwap encounters difficulties in extracting facial features from unevenly lit source faces, resulting in blurred results. And RAFSwap generates some incorrect transferings (note the red box markings). Our FlowFace++ is still capable of generating high-quality face-swapping results even in the presence of uneven illumination. 
% 原因分析
% 0、mae提取特征的能力
% 1、CAFM融合特征的能力
% 2、exp loss

Our FlowFace++ achieves remarkable performance in robustness testing, primarily because we utilize the MAE encoder which is designed following MAE and pre-trained on a large-scale face dataset using the masked training strategy. Even under extreme input conditions with various interferences, the MAE encoder is able to extract rich features from the input faces. 
Furthermore, our CAFM module facilitates the adaptive aggregation of identity and facial shape information from the source, allowing us to effectively eliminate the interfering information.
% Furthermore, thanks to the adaptive aggregation of identity and facial shape information from the source using the CAFM module, interfering information can be dynamically removed. 

% exp-loss需要进一步补充
To maintain the expression in the generated image consistent with the target image, we introduce an effective expression embedding~\cite{zhang2021learning} which employs a continuous and compact embedding space to represent the fine-grained expressions. The distance between two expression embedding reflects the similarity between them. Therefore, we formulate the expression loss as a L2 loss that computes the expression embedding distance between the generated and target images, so as to ensure that they are consistent in terms of expression.

% \vspace{-0.2cm}
\subsection{\textbf{Analysis of FlowFace++}}

Three ablation studies are conducted to validate our end-to-end FlowFace++ framework and several components used in $D^{shape}$ and $F^{swa}$, respectively.

\begin{figure}[t]
\centering
\includegraphics[width=0.48\textwidth]{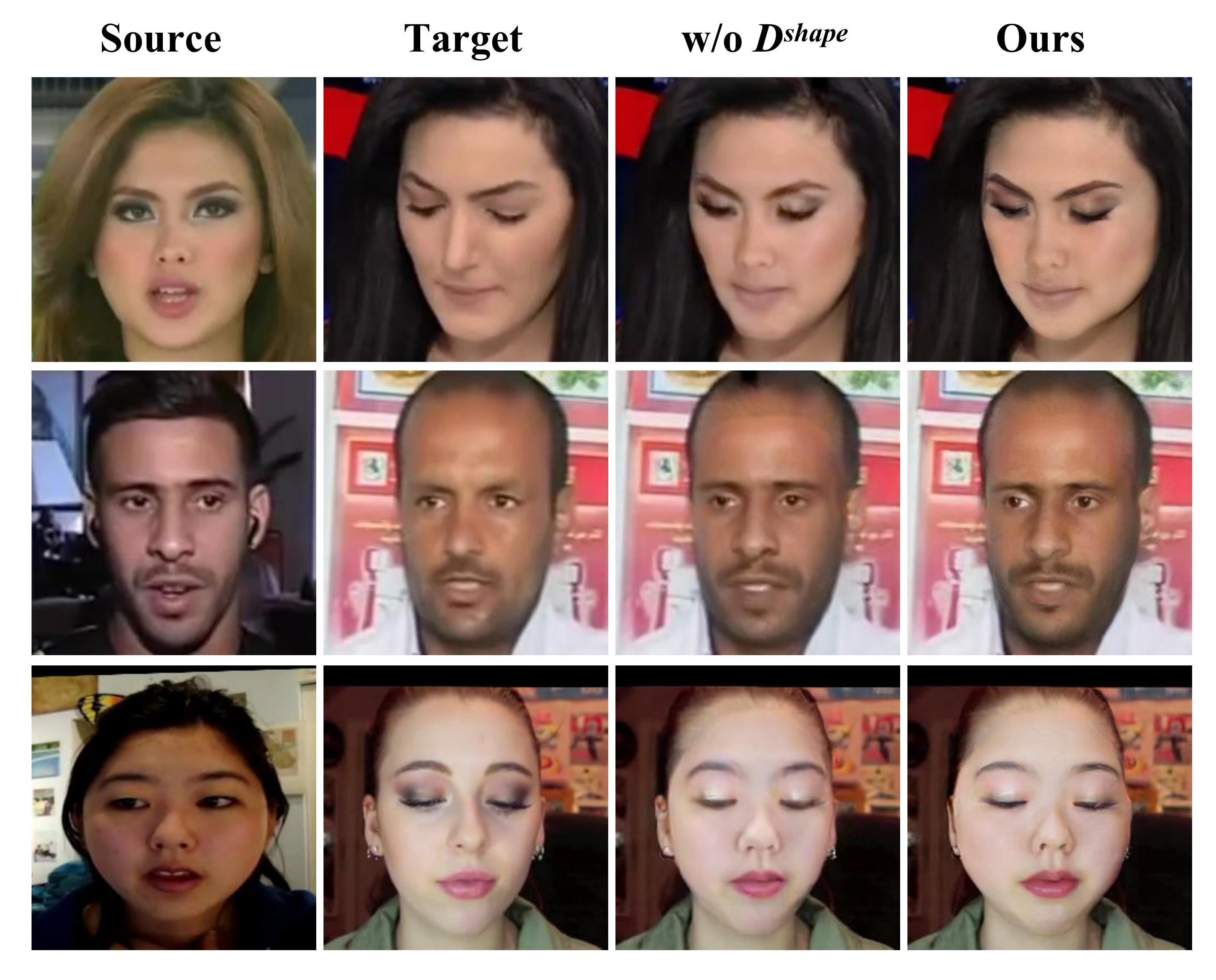}
\caption{
Qualitative ablation results of $D^{shape}$.
}
\label{fig:compare_noFlow}
\end{figure}

\subsubsection{Ablation Study on FlowFace++.} 

We conduct ablation experiments to validate the design of end-to-end face-swapping framework $F^{swa}$. We remove the $D^{shape}$ (w/o $D^{shape}$). Figure~\ref{fig:compare_noFlow} shows that in the absence of $D^{shape}$'s constraints, the network lacks the ability to transfer facial shape (note the cheeks and jaw angles). 
Hence, $D^{shape}$'s constraints are crucial in ensuring our FlowFace++'s proficiency in facial shape warping.

\begin{figure}[t]
\centering
\includegraphics[width=0.48\textwidth]{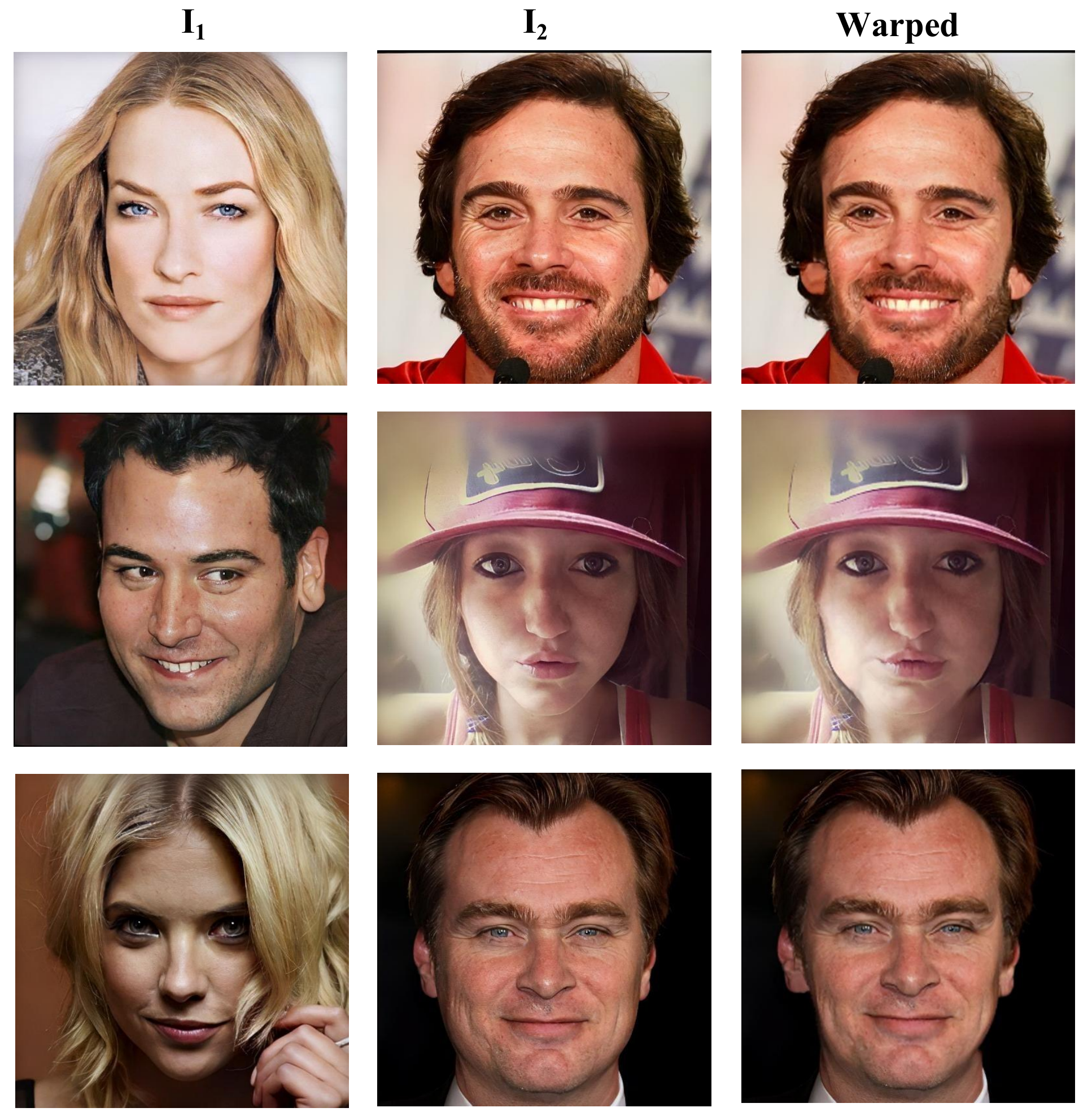}
\caption{
Qualitative ablation results of $D^{shape}$. After warping the input faces with semantic flows, the facial contours are altered while maintaining the non-shape identity information.
}
\label{fig:ablation_onlyDshape}
\end{figure}

To further validate the effectiveness of $D^{shape}$, we attempt to warp the input faces using the semantic flows generated by $D^{shape}$. As shown in Figure~\ref{fig:ablation_onlyDshape}, after warping with the semantic flow, the facial contours of \textbf{$I_2$} are changed, becoming closer to \textbf{$I_1$}, while maintaining the inner-facial appearances. This fully demonstrates that the semantic flow generated by our designed $D^{shape}$ effectively represents the pixel-wise motions in facial contours explicitly.

Our previous work, FlowFace\cite{zeng2022flowface}, is a two-stage framework. It uses the $D^{shape}$'s warped results on the target faces as the first stage, then utilizes the $F_{swa}$ which is trained without $D^{shape}$'s supervision to transfer the non-shape identity as the second stage. 

% Furthermore, we modify FlowFace++ into a two-stage framework and design another comparative experiment. We use the $D^{shape}$'s warped results on the target faces as the first stage of this new framework, then train a FlowFace++ network without adding $D^{shape}$ supervision, to transfer the identity of the source faces to the warped target faces. This is used as the second stage of the new framework. 

\begin{figure}[t]
\centering
\includegraphics[width=0.48\textwidth]{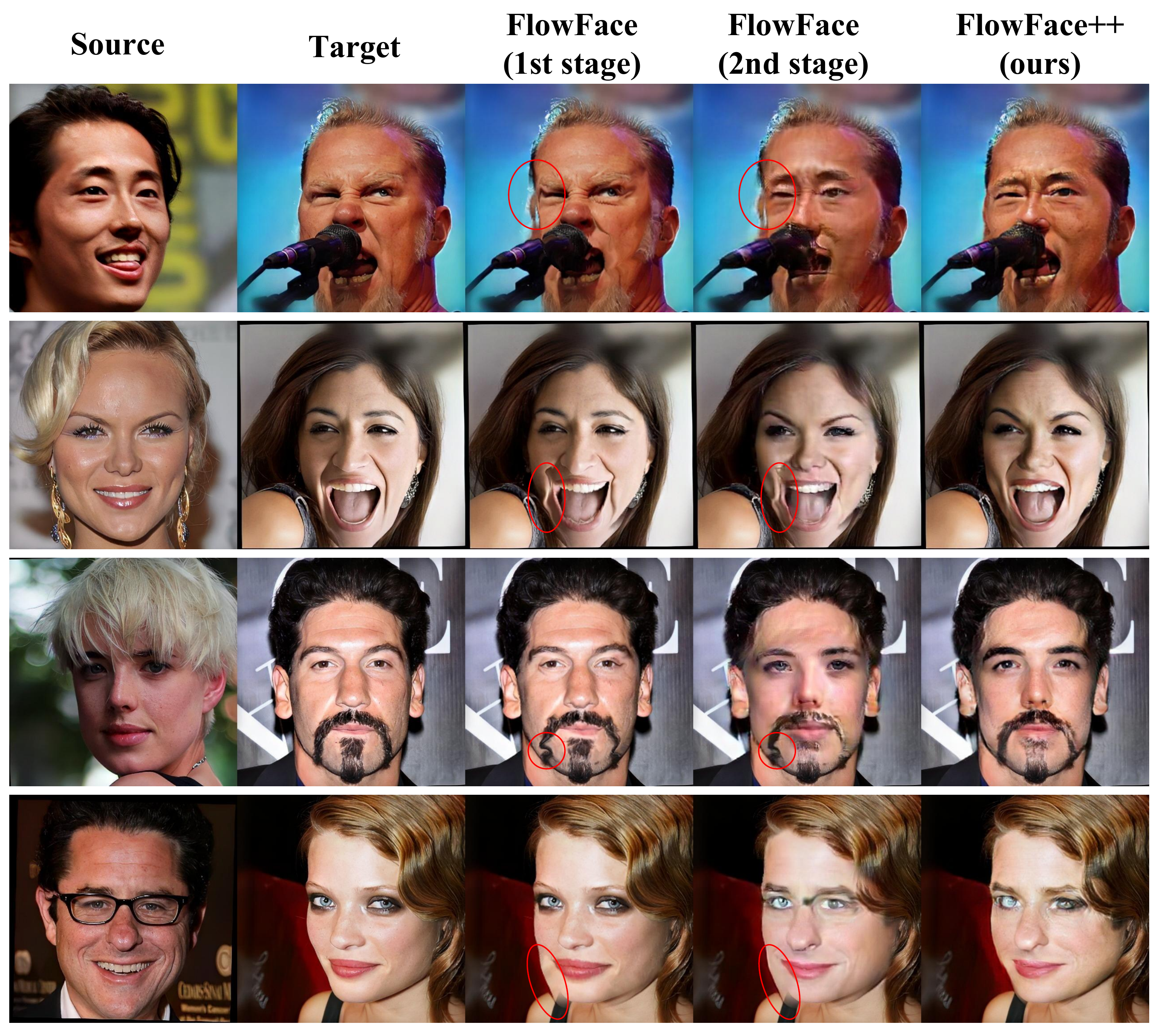}
\caption{
Qualitative ablation results of our previous framework FlowFace. In the first stage of FlowFace, the target faces undergo warping via $D^{shape}$. In the second stage of FlowFace, the non-shape identity of the source faces is transferred to the warped target faces.
}
\label{fig:ablation_twostage}
\end{figure}

As shown in Figure~\ref{fig:ablation_twostage}, the performance of FlowFace\cite{zeng2022flowface} on the target faces via $D^{shape}$ is not entirely perfect and may result in distortions in details (\emph{e.g.}, eyebrow, mouth and low jawbone), as highlighted by the red circle in the image. Such distortions may emerge in the second stage, potentially leading to flaws in the final results. This convincingly underscores the rationale behind our end-to-end network architecture of FlowFace++, which incorporates $D^{shape}$ as a form of supervision.

\begin{figure*}[t]
\centering
\includegraphics[width=0.9\textwidth]{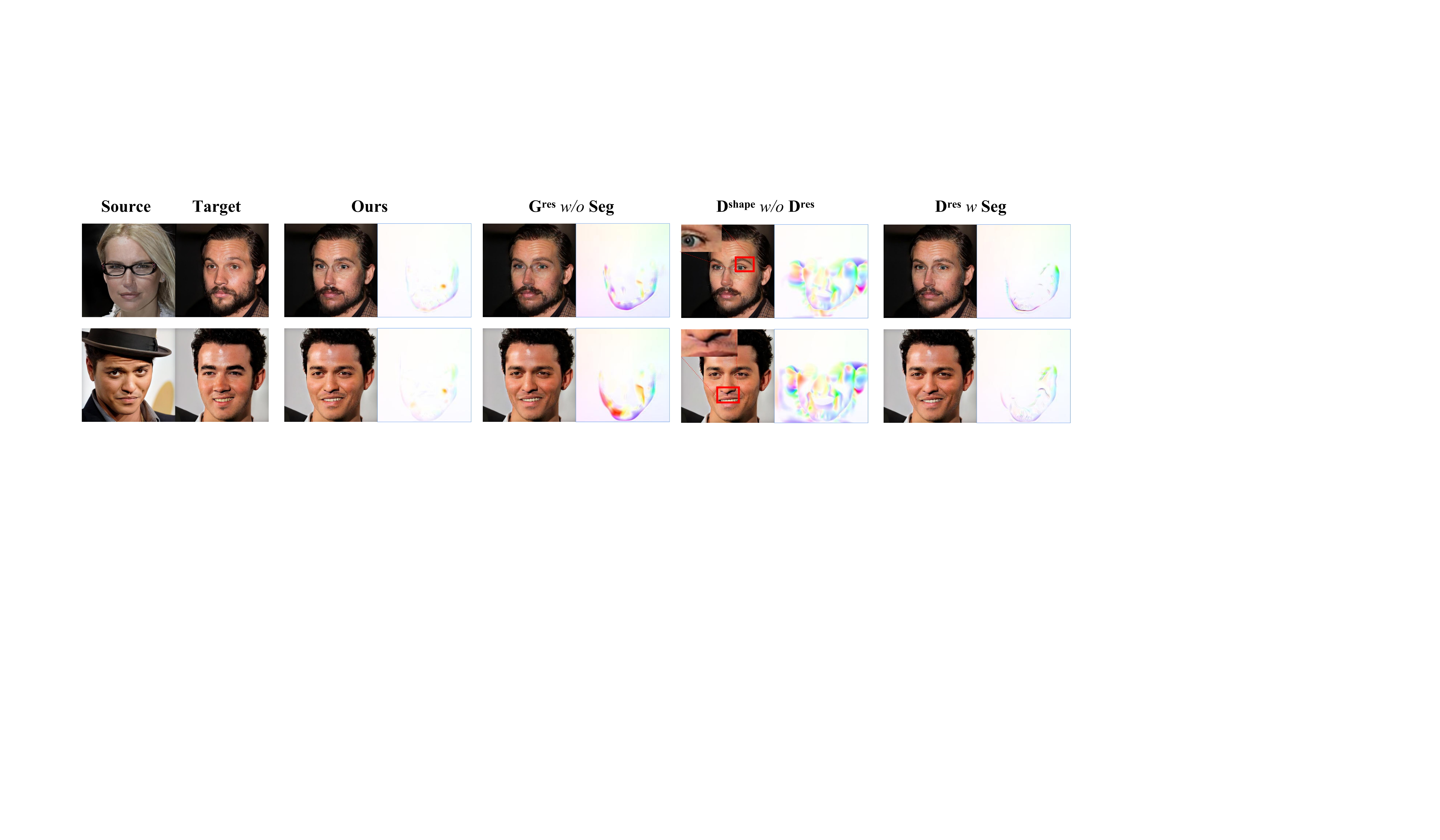}  
\caption{Qualitative ablation results of each component in $D^{shape}$.}
\label{fig:ablation_Dshape}
\end{figure*}

\begin{table}[h]
% \vspace{-0.1in}
\setlength{\abovecaptionskip}{0.2cm}
\centering
% \caption{Quantitative ablation study of $F^{swa}$ on FF++.}
\resizebox{0.5\textwidth}{!}{
\begin{tabular}{c|ccc|c|c|c}
\multirow{2}{*}{Methods} & \multicolumn{3}{c|}{ID Acc($\%$) $\uparrow$}                     & \multicolumn{1}{l|}{\multirow{2}{*}{Shape$\downarrow$}} & \multicolumn{1}{l|}{\multirow{2}{*}{Expr.$\downarrow$}} & \multicolumn{1}{l}{\multirow{2}{*}{Pose.$\downarrow$}} \\ \cline{2-4}
                         & CosFace        & SphereFace     & Avg            & \multicolumn{1}{l|}{}                       & \multicolumn{1}{l}{}    
                         & \multicolumn{1}{l}{}      
                         \\ \hline
\textit{Sparse Ldmks}                 & 99.01     & 98.16 & 98.59 & 1.49 &   0.23 & 1.86                                        \\
\textit{$D^{shape}$w/o $D^{res}$}                 & 98.43          & 97.01 & 97.72 & 1.55 & \textbf{0.22}      &   1.78                                        \\
\textit{$D^{res}$w Seg}                 & 98.72     & 97.34 & 98.03 & 1.53 &   0.28 &1.91                                        \\
\textit{$G^{res}$w/o Seg}                 & 96.79     & 96.09 & 96.44 & 1.57 &   \textbf{0.22} &\textbf{1.74}                                        \\

\hline
\textit{$Ours$}                 & \textbf{99.51}     & \textbf{99.03} & \textbf{99.27} & \textbf{1.43} &   0.23 &2.20                       \\                                    
\end{tabular}
}
\caption{Quantitative ablation study of $D^{shape}$ on FF++.}
% \vspace{-1em}
\label{tab:ablation_Dshape}
\end{table}

% \vspace{-0.2cm}
\subsubsection{Ablation study on $D^{shape}$.}
We design four ablation experiments to validate the effectiveness of our proposed $D^{shape}$:

\textit{A. Sparse Landmarks vs. Dense Flow. }During the training process of $F_{swa}$, we adopt the dense flow $V_so$ generated by $D_{shape}$ to calculate the shape discrepancies between $I_s$ and $I_o$. To demonstrate the rationality of dense flow, similar to the approach of HifiFace\cite{wang2021hififace}, we replace it with sparse landmarks as the supervision for facial shape, where the sparse landmarks $P_{12}$ can be obtained by calculating Equation~\ref{con:landmarks}. Then the new facial shape loss can be computed by:
\begin{equation}
\mathcal{L}_{sparse\_ldmks} = \left\|P_{12} - P_{o}\right\|_{2},\
\end{equation}
The $P_o$ represents the landmarks of the $I_o$ and only 17 landmarks on the contour are involved in the calculation of ${L}_{sparse\_ldmks}$. As seen in Figure~\ref{fig:ablation_Dshape}, when sparse landmarks are used as supervision, residual ghosts may appear on the facial contour during the process of transferring facial shape. This phenomenon can be attributed to that sparse landmarks cannot represent pixel-wise dense motion.

\textit{B. Removing $S_2$ of $G^{res}$. }We Remove the semantic input $S_2$ of $G^{res}$ ($G^{res}$ w/o Seg) to validate our proposed semantic guided generator $G^{res}$. It can be seen from Figure~\ref{fig:ablation_Dshape} that some inaccurate flow occurs in the generated face, which implies that only facial landmarks cannot guide $G^{res}$ to produce accurate dense flow due to the lack of semantic information. 
The results also demonstrate that the semantic information is beneficial for accurate flow estimation and validates $G^{res}$.

% Specifically, we remove the semantic input $S_t$ of $G^{res}$ ($G^{res}$ w/o Seg). It can be seen from Figure~\ref{fig:ablation_Dshape} that some inaccurate flow occurs in the generated face, which implies that only facial landmarks cannot guide $G^{res}$ to produce accurate dense flow due to the lack of semantic information. 
% The results also demonstrate that the semantic information is beneficial for accurate flow estimation and validates $G^{res}$.
\textit{C. Adding the semantic inputs ($S_2$ and $S_2^{res}$) of $D^{res}$ ($D^{res}$ w Seg). } FlowFace \cite{zeng2022flowface} proposes that adding semantic inputs to the adversarial loss can improve the reconstruction of fine details in the facial region. We attempt to add semantic inputs to $D^{res}$ as well, however, as shown in the experimental results in Table~\ref{tab:ablation_Dshape} suggest that this does not result in any significant accuracy improvements. The implication is that in FlowFace++, $D^{res}$ does not require semantic inputs and is still capable of effectively discriminating between real and generated faces. 
% This means that in FlowFace++, image inputs alone are sufficient to discern real and generated faces in $D^{res}$.

\textit{D. Removing $D^{res}$ (w/o $D^{res}$). }As observed in Figure~\ref{fig:ablation_Dshape}, the generated images produced by w/o $D^{res}$ exhibit a few artifacts, and the estimated semantic flow is also marred by substantial noise.

The aforementioned observations serve to corroborate the efficacy of our proposed $D^{shape}$.

% Then, we conduct two ablation experiments to validate $D^{res}$: 

% (1) Adding the semantic inputs ($S_t$ and $S_t^{res}$) of $D^{res}$ ($D^{res}$ w Seg). 
% In FlowFace, the authors propose that adding semantic inputs to the adversarial loss can improve the reconstruction of fine details in the facial region. We attempt to add semantic inputs to $D^{res}$ as well, however, as shown in The experimental results in Table~\ref{tab:ablation_Dshape} suggest that this does not result in any significant accuracy improvements.  We believe that image inputs alone are sufficient to discern real and generated faces in $D^{res}$.

% (2) removing $D^{res}$ (w/o $D^{res}$). As observed in Figure~\ref{fig:ablation_Dshape}, the generated images produced by w/o $D^{res}$ exhibit a few artifacts, and the estimated flow is also marred by substantial noise. 
% The aforementioned observation serves to corroborate the efficacy of our proposed $D^{shape}$.

\begin{figure}[t]
\setlength{\abovecaptionskip}{0.16cm}
\centering
\includegraphics[width=0.46\textwidth]{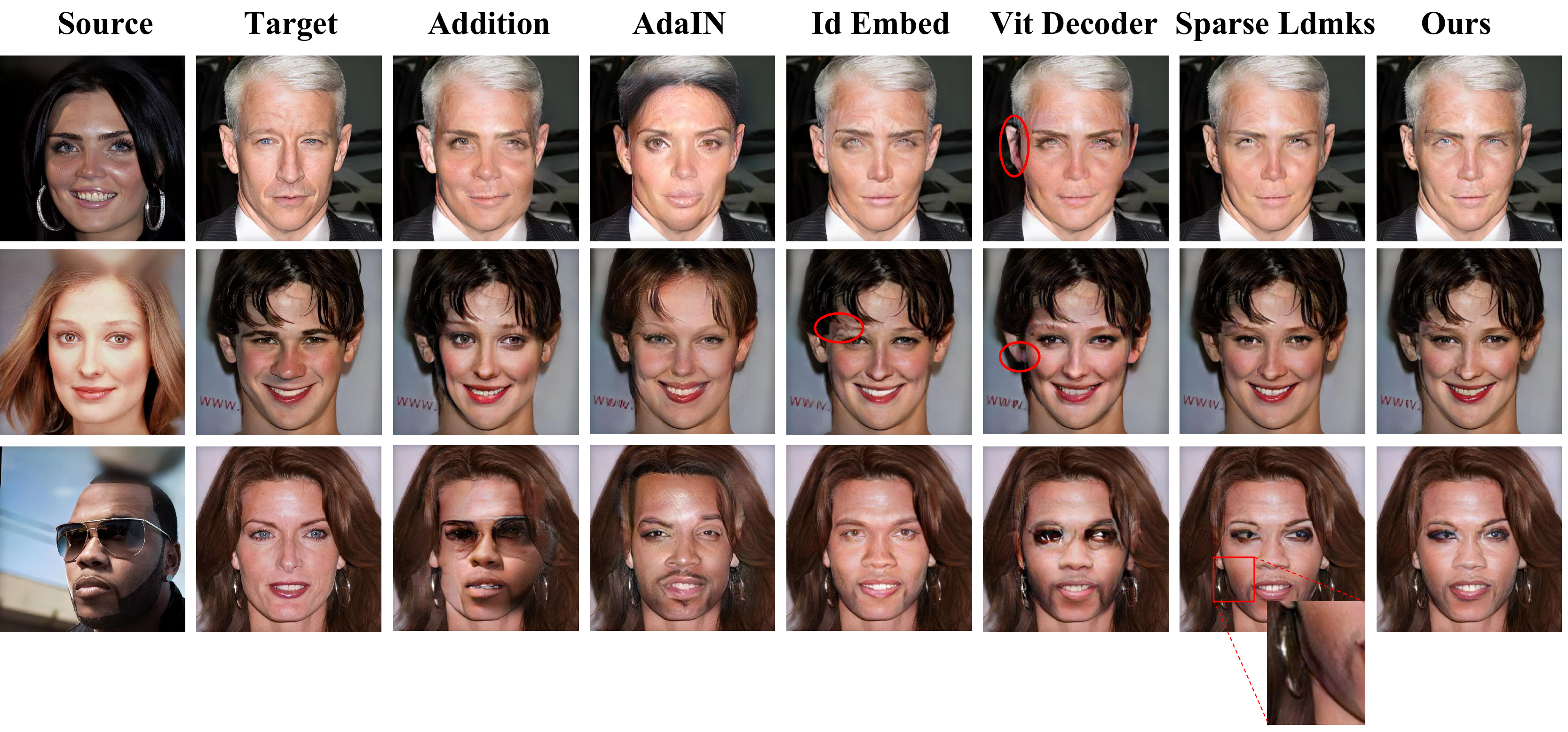}  
\caption{Qualitative ablation study of $F^{swa}$.}
\label{fig:ablation_second}
\end{figure}

\begin{figure}[t]
\setlength{\abovecaptionskip}{0.16cm}
\centering
\includegraphics[width=0.46\textwidth]{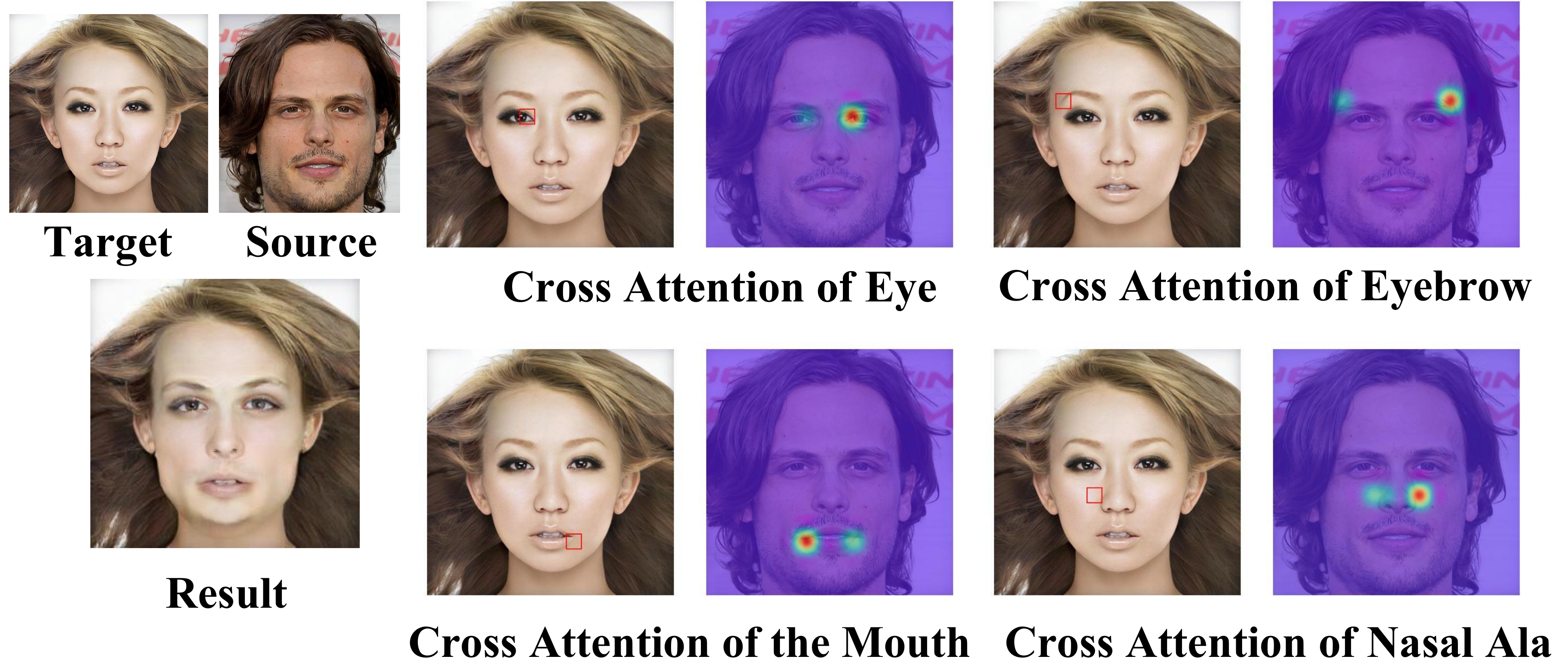}  
\caption{Visualize the cross-attention of different facial parts. For each part in the target, our CAFM can accurately focus on the corresponding parts in the source.
}
% \vspace{-1em}
\label{fig:cross_att_vis}
\end{figure}

\begin{table}[t]

% \vspace{-0.1in}
\setlength{\abovecaptionskip}{0.2cm}
\centering
% \caption{Quantitative ablation study of $F^{swa}$ on FF++.}
\resizebox{0.45\textwidth}{!}{
\begin{tabular}{l|ccc|c|c|c}
\multirow{2}{*}{Methods} & \multicolumn{3}{c|}{ID Acc}                     & \multicolumn{1}{l|}{\multirow{2}{*}{Expr}} & \multicolumn{1}{l}{\multirow{2}{*}{Pose}} & \multicolumn{1}{l}{\multirow{2}{*}{Shape}} \\ \cline{2-4}
                         & CosFace        & SphereFace     & Avg            & \multicolumn{1}{l|}{}                       & \multicolumn{1}{l}{}                       \\ \hline
\textit{Addition}                 & 99.11          & \textbf{99.31} & \underline{99.21} & 0.35                                        &   4.29            &   1.30                            \\
\textit{AdaIN}                 & 33.53          & 25.62 & 29.58 & 0.29                                        &   2.67     &2.63                                   \\
\hline
\textit{Id Embed.}                   & 96.76          & 95.44          & 96.10          & \textbf{0.23}                                  & \textbf{1.78}    & 1.71                                      \\
\textit{Vit}                & \underline{99.42}          & 98.85          & 99.14          & 0.26                                       & 2.69        &\textbf{1.39}                                 \\
\textit{Ours}                     & \textbf{99.51} & \underline{99.03}    & \textbf{99.27}    & \textbf{0.23}                               & \underline{2.20}       &\underline{1.43}                               
\end{tabular}
}
\caption{Quantitative ablation study of $F^{swa}$ on FF++.}
% \vspace{-1em}
\label{tab:ablation_swap}
\end{table}

% \vspace{-0.2cm}
\subsubsection{Ablation study on $F^{swa}$.}
% Figure~\ref{fig:ablation_second} and Table~\ref{tab:ablation_swap} 
Three ablation experiments are conducted to evaluate the design of $F^{swa}$:

\textit{A. Choices on CAFM, \textit{Addition} and \textit{AdaIN}.} To verify the effectiveness of our proposed CAFM, we conduct a comparative analysis between the CAFM and two other methods, namely, \textit{Addition} (which simply involves adding the source values to the target values) and \textit{AdaIN} (which first averages source patch embeddings and then injects them into the target feature map using AdaIN residual blocks).
As shown in Figure~\ref{fig:ablation_second} and Table~\ref{tab:ablation_swap}, \textit{Addition} simply blends all information of the source face to the target face, thus resulting in severe pose and expression mismatch. \textit{AdaIN}, due to its global modulation, impacts not only the facial features but also the non-face parts such as hair. 
By contrast, $F^{swa}$ with CAFM achieves a high ID Acc and effectively preserves the target attribute. This result demonstrates the highly accurate identity information extraction capabilities of CAFM from the source face and its adaptive infusion into the target counterpart.
% In contrast, $F^{swa}$ with CAFM obtains a high ID Acc and preserves the target attribute well, which proves that CAFM can accurately extract identity information from the source face and adaptively infuse it into the target counterpart. 

To further validate the efficacy of our CAFM, we visualize the cross-attention computed by CAFM. As depicted in Figure~\ref{fig:cross_att_vis}, when a specific part of the target face is given (marked with red boxes), our proposed CAFM precisely focuses on the corresponding regions of the source face, thus validating its ability to adaptively transfer the identity information from the source patches to the respective target patches.
% As shown in Figure~\ref{fig:cross_att_vis}, given a specific part (marked by red boxes) of the target face, CAFM accurately focuses on the corresponding parts of the source face, validating our CAFM can adaptively transfer the identity information from the source patches to corresponding target patches.

\textit{B. Latent Representation vs. ID Embedding (\textit{ID Embed.}). } 
To verify the superiority of using the latent representation of MAE, We design an experiment that use a face recognition network~\cite{arcface_irse} to extract the features of the source faces and continue to use the original MAE encoder to extract the features of the target faces, while keeping the other network structures unchanged.
% we train a new model which adopts the identity embedding as the identity representation and employs AdaIN as the injection method.
As can be seen from Figure~\ref{fig:ablation_second}, \textit{ID Embed.} misses some fine-grained face appearances, such as eyebrow edge. 
In contrast, $F^{swa}$ contains richer identity information and achieves higher ID Acc, as shown in Tab~\ref{tab:ablation_swap}.

\textit{C. Convolutional Decoder vs. ViT Decoder. }
% (\textit{ViT}). 
We try two different decoders to determine the better one.
As shown in Table~\ref{tab:ablation_swap}, compared to \textit{Vit Decoder}, \textit{Convolutional Decoder} exhibits superior performance in terms of ID accuracy, expression error, and pose error, while performing roughly on par in the aspect of shape error. As can be seen in Figure~\ref{fig:ablation_second}, the result of \textit{Vit Decoder} exhibits partial blurry regions and erroneous leakage of the source face's hair. 

\section{Conclusion}
This work proposes a novel face-swapping framework, FlowFace++, which utilizes explicit semantic flow supervision and an end-to-end architecture to facilitate shape-aware face swapping. Specifically, our work pretrains a shape-aware discriminator to supervise the face swapping network thus optimizing it to generate highly realistic results. The face swapping network is a stack of a pre-trained face-masked autoencoder (MAE), a cross-attention fusion module, and a convolutional decoder. MAE is used to extract facial features that better capture facial appearances and identity information. The cross-attention fusion module is designed to better fuse the source and the target features, thus leading to better identity preservation. 

We conduct extensive quantitative and qualitative experiments on in-the-wild faces, demonstrating that FlowFace++ outperforms the state-of-the-art significantly. In the quantitative experiments, our FlowFace++ demonstrates effective performance with four metrics, including identity retrieval accuracy, shape error, expression error, and pose error. In the qualitative experiments, our approach achieves higher similarity with the source faces in terms of facial shape and inner facial features. Furthermore, we compare the performance of our FlowFace++ with other methods under extreme input conditions, and it exhibits higher stability when handling source faces with large angular deviations or non-uniform illumination.

We further conduct comprehensive ablation experiments to validate the rationality of the FlowFace++ design. The experimental results demonstrate the irreplaceable superiority of our facial shape discriminator, MAE encoder, cross-attention fusion module, and convolutional decoder in achieving high-quality transferring of inner-facial appearances and facial shape. 
% Extensive quantitative and qualitative experiments are conducted on in-the-wild faces, demonstrating that our FlowFace++ outperforms the state-ofthe-art significantly.

Despite the superior performance of our FlowFace++, there are still some limitations. Our FlowFace++ still faces the challenge of lacking temporal constraints, which has not been explicitly addressed by previous face-swapping methods. Besides, in situations where the source face is wearing sunglasses, it may cause distortion in the eye regions of the resultant face.

\begin{figure*}[t]
    \centering
    \includegraphics[width=0.9\textwidth]{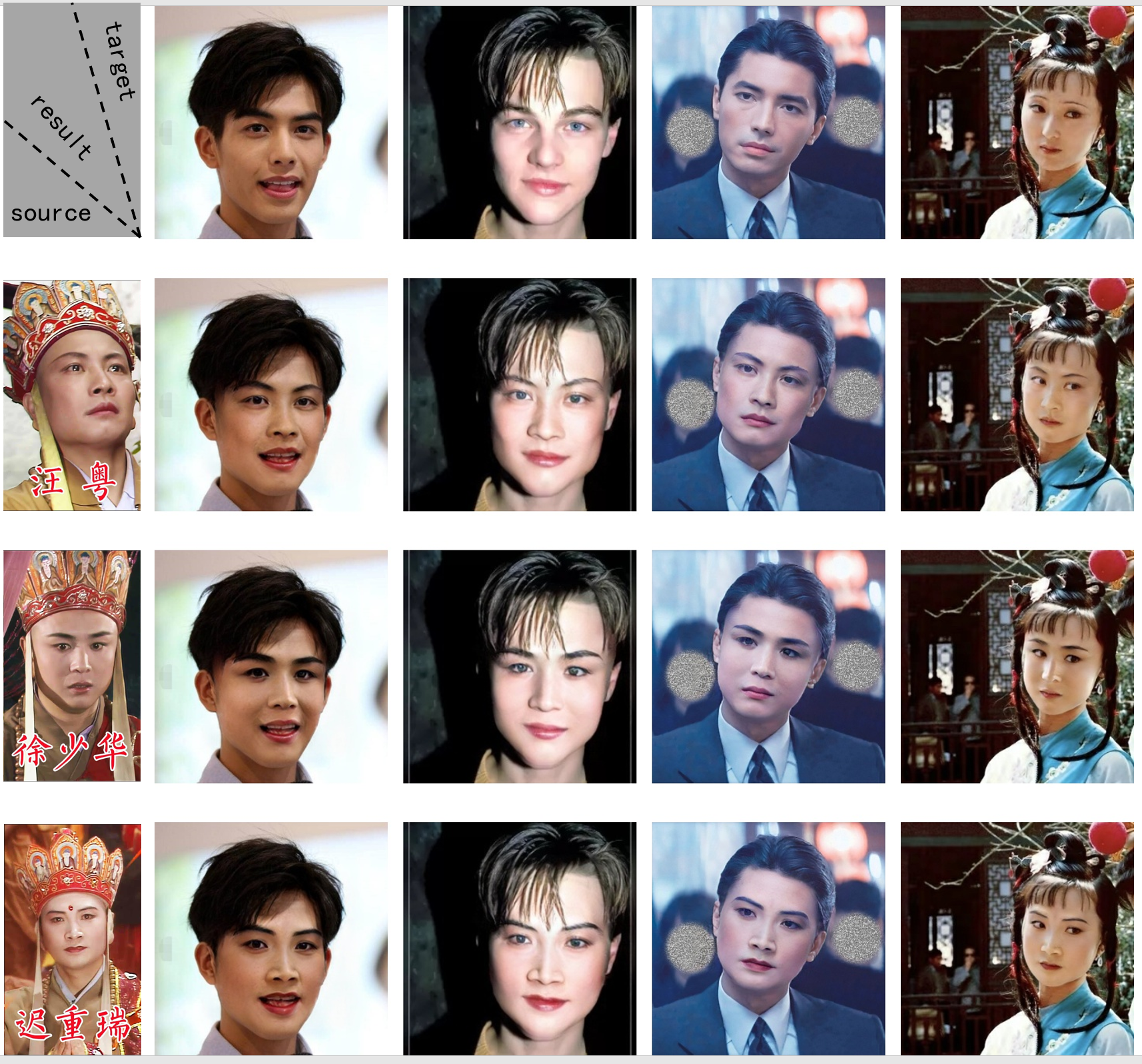}  
    \caption{More face-swapping results generated by our FlowFace++.}
    \label{fig:ablation_Dshape}
\end{figure*}

%Therefore, when applied to scenarios where continuous video frames are used as input, the resultant video may suffer from flicker.

% \section{Limitation}
% Despite the superior performance of our FlowFace++ over current state-of-the-art methods, there are still some limitations. 
% The current face swapping methods generally lack temporal constraints, and our FlowFace++ is no exception. Therefore, when applied to scenarios where continuous video frames are used as input, the resultant video may suffer from flicker.
% (jittering of the background or facial regions).

% \bibliography{egbib}
\bibliographystyle{IEEEtran}
\bibliography{main}

\vfill

\end{document}